\documentclass[final,5p,times, sort&compress, authoryear]{elsarticle}

\usepackage{amssymb}
\usepackage[cmex10]{amsmath}
\usepackage{multicol}
\usepackage{multirow}
\usepackage{tikz}
\usetikzlibrary{shapes,arrows,fit, calc, backgrounds, positioning}
\tikzstyle{block} = [draw, rectangle, text width=2.7cm, fill=gray!20,  text centered, minimum height=1.2cm, node distance=2.5cm, rounded corners=.25ex]
\usepackage{array}
\usepackage[colorlinks]{hyperref}
\usepackage[]{subcaption}
\usepackage{booktabs}
\newcommand{\toppad}{12pt}
\newcommand{\botpad}{6pt}
\newcommand\setrow[1]{\gdef\rowmac{#1}#1\ignorespaces}
\newcommand\clearrow{\global\let\rowmac\relax}
\clearrow

\journal{Expert Systems with Applications}

\begin{document}

\begin{frontmatter}



\title{Video-based Sequential Bayesian Homography Estimation for Soccer Field Registration}


\author[up]{Paul Johannes Claasen\corref{cor1}}
\ead{pj@benjamin.ng.org.za}

\author[up]{Johan Pieter de Villiers}
\ead{pieter.devilliers@up.ac.za}

\affiliation[up]{organization={Department of Electrical, Electronic and Computer Engineering, University of Pretoria},
            addressline={Lynnwood Road, Hatfield}, 
            city={Pretoria},
            postcode={0028}, 
            country={South Africa}}

\cortext[cor1]{Corresponding author.}

\begin{abstract}
A novel Bayesian framework is proposed, which explicitly relates the homography of one video frame to the next through an affine transformation while explicitly modelling keypoint uncertainty. The literature has previously used differential homography between subsequent frames, but not in a Bayesian setting. In cases where Bayesian methods have been applied, camera motion is not adequately modelled, and keypoints are treated as deterministic. The proposed method, Bayesian Homography Inference from Tracked Keypoints (BHITK), employs a two-stage Kalman filter and significantly improves existing methods. Existing keypoint detection methods may be easily augmented with BHITK. It enables less sophisticated and less computationally expensive methods to outperform the state-of-the-art approaches in most homography evaluation metrics. Furthermore, the homography annotations of the WorldCup and TS-WorldCup datasets have been refined using a custom homography annotation tool that has been released for public use. The refined datasets are consolidated and released as the consolidated and refined WorldCup (CARWC) dataset.
\end{abstract}



\begin{keyword}
    homography estimation \sep sports field registration \sep camera calibration \sep keypoint detection \sep monocular vision


\end{keyword}

\end{frontmatter}


\section{Introduction}
\label{sec:intro}
    \sloppy Homography estimation generally refers to a planar projective transformation that relates corresponding points in two views of the same scene. Homography estimation plays a crucial role in many computer vision applications. Examples include automated panoramic image stitching \citep{pano}, simultaneous localisation and mapping (SLAM) \citep{orb_slam, deep_homo, deep_homo_dynamic}, camera calibration and pose estimation \citep{camera_calib, real_time_pose} and video stabilisation \citep{video_stab}. Sports field registration applies specifically to the case where one of the scene views represents a structured model of a sports field. Sports field registration enables aligning virtual overlays, such as graphics, annotations, or analysis tools, with the real-world sports field, as well as accurate player tracking and augmented reality experiences. 
    
    While the application of interest in this work is sports field registration, the Bayesian homography estimation method presented in this paper could also be applied to some of these other applications under certain conditions. For example, given a suitable keypoint matching algorithm and reasonable estimates of the process and measurement noise parameters, the method could be applied to estimate the homography of planar scenes in SLAM applications \citep{orb_slam}. It may also be employed to estimate camera pose as in \citep{real_time_pose}. While this paper focuses on the specific use case of soccer field registration, it is hoped to inspire similar approaches in these application areas.

    \subsection{Contributions}
        The main contributions of this paper are as follows:
            \begin{itemize}
                \item Whereas previous methods treat keypoints used for homography estimation as deterministic, this work considers them stochastic.
                \item A novel dynamics model is proposed which explicitly relates the homography of immediately subsequent frames to one another.
                \item A two-stage Kalman filter approach is used to filter the homography from tracked keypoints.
                \item The proposed approach improves the results of a relatively simple keypoint detector to outperform an expensive state-of-the-art method.
                \item Finally, the WorldCup \citep{deep_structured} and TS-WorldCup \citep{kpsfr} datasets have been refined using a custom homography annotation tool (released for public use) and released as the consolidated and refined WorldCup (CARWC) dataset.
            \end{itemize}

        The remainder of the article is structured as follows. Section 2 provides the necessary background of projective spaces to understand what the homography matrix represents. Section 3 provides an overview of previous works in the literature concerning homography estimation. Section 4 describes the proposed approach in detail, including mathematical derivations, and summarises the main differences between the proposed approach and previous methods from the literature. Section 5 describes the experiments used to evaluate the proposed method. This section includes parameter settings, details regarding datasets, a description of how covariance matrices were measured and examples of their values and formulae for the evaluation metrics that were used. The results of these evaluations are presented and discussed in Section 6. Finally, the main conclusions are presented in Section 7.

    \section{Background}
    \label{sec:background}
        The world in $\mathbb{R}^3$ is imaged through a projective camera, resulting in a 3D projective space $\mathbb{P}^3$, which augments $\mathbb{R}^3$ with points at infinity \citep{multiple_view_geometry}. A coordinate $\mathbf{X} = \begin{bmatrix}
            X & Y & Z
        \end{bmatrix}$ in $\mathbb{R}^3$ is augmented to form $\mathbf{X}' = \begin{bmatrix}
            X & Y & Z & T
        \end{bmatrix}^\top$ (termed a homogeneous coordinate) in $\mathbb{P}^3$, with the corresponding point at infinity occurring for $T=0$. Although the 3D vector is augmented by the element $T$, the projective space is, by convention, still considered three-dimensional. Hence, the superscript of $\mathbb{P}$ remains 3. A projective camera then performs a linear mapping on the homogeneous coordinate $\mathbf{X}'$ from the 3D projective space $\mathbb{P}^3$ (which represents the world space) to the homogeneous coordinate $\mathbf{x}' = \begin{bmatrix}
            x & y & t
        \end{bmatrix}^\top$ in the 2D projective space $\mathbb{P}^2$ (which represents the image space). The transformation from 3D to 2D projective space is governed by \begin{equation*}
            \begin{bmatrix}
                x \\ y \\ t
            \end{bmatrix} = \mathbf{P}\begin{bmatrix}
                X \\ Y \\ Z \\ T
            \end{bmatrix},
        \end{equation*} where $\mathbf{P}=\boldsymbol{\kappa}\mathbf{\left[R|C\right]}$ and $|$ denotes column-appended matrix augmentation. $\boldsymbol{\kappa}\in \mathbb{R}^{3\times3}$ represents the internal camera parameters and has the general form \begin{equation*}
                \boldsymbol{\kappa} = \begin{bmatrix}
                    \alpha_x & s & x_0\\
                    0 & \alpha_y & y_0\\
                    0 & 0 & 1
                \end{bmatrix},
            \end{equation*} 
        where $\alpha_x$ and $\alpha_y$ are scale factors in the x- and y-coordinate directions, respectively, $s$ is the skew which is non-zero if the x- and y-axes are not perpendicular and $\begin{bmatrix}
            x_0 & y_0
        \end{bmatrix}^\top$ represents the coordinates of the principal point -- the geometric centre of the image. For further details regarding $\alpha_x$, $\alpha_y$, $s$, $x_0$ and $y_0$, the reader is referred to \citep{multiple_view_geometry}. $\mathbf{R}\in \mathbb{R}^{3\times3}$ and $\mathbf{C}\in \mathbb{R}^3$ respectively relate the camera orientation (rotation) and position (translation) to the world coordinate system. These are the external camera parameters \citep{multiple_view_geometry}. An arbitrary homogeneous vector $\mathbf{x}' = \begin{bmatrix}
            x & y & t
        \end{bmatrix}^\top$ in $\mathbb{P}^2$ may be normalised to become $\begin{bmatrix}
            x/t & y/t & 1
        \end{bmatrix}^\top$, $t\neq 0$, which represents \begin{equation}
            \label{normalisation}
                \begin{bmatrix}
                    x/t & y/t
                \end{bmatrix}^\top
            \end{equation}
        in $\mathbb{R}^2$, a point in the image  \citep{multiple_view_geometry}. That is, for a constant $\mathbf{x}'$, $k\mathbf{x}'$ represents the same point in the image for any $k\neq 0$ and may be thought of as a ray in $\mathbb{R}^3$ passing through the centre of projection of the camera. One may consider the projective space $\mathbb{P}^2$ as a space consisting of the set of such rays, each representing a single point in the image in $\mathbb{R}^2$. Finally, if all the world points are coplanar such that $Z=0$, the transformation from 3D to 2D projective space is performed by the homography matrix $\mathbf{H}$: \begin{equation}
            \label{homo_mapping}
            \begin{bmatrix}
                x \\ y \\ t
            \end{bmatrix} = \mathbf{H}\begin{bmatrix}
                X \\ Y \\ T
            \end{bmatrix} = \begin{bmatrix}
                h_{11} & h_{12} & h_{13} \\
                h_{21} & h_{22} & h_{23} \\
                h_{31} & h_{32} & h_{33} \\
            \end{bmatrix}\begin{bmatrix}
                X \\ Y \\ T
            \end{bmatrix}.
        \end{equation} 
        It should be noted that $\mathbf{H}$ is determined up to a scale and thus has only eight degrees of freedom \citep{multiple_view_geometry}.

\section{Related work}

This section examines previous homography estimation methods from the literature to illustrate the novelty of the proposed method.

\label{sec:related_work}

    \subsection{Homography estimation}
    \label{sec:homography_estimation}
    Homography estimation is typically achieved by identifying corresponding features, or keypoints, between two images. This may be achieved through the use of features extracted by methods such as the scale-invariant feature transform (SIFT) \citep{sift} or ORB (Oriented FAST and Rotated BRIEF, where FAST refers to a keypoint detection method and BRIEF refers to another feature descriptor \citep{orb}), and matching methods such as k-nearest neighbours \citep{pano} or bags of words \citep{orb_slam}. These keypoints are subsequently used to estimate the mapping between the images with the direct linear transform (DLT) \citep{multiple_view_geometry} or random sample consensus (RANSAC) \citep{ransac} algorithms. Another approach to homography estimation is that of iterative optimisation such that the alignment between a target image and the transformation of another image is maximised through the minimisation of a chosen loss function \citep{lucas, parametric}. These methods are usually slower than feature-based methods. Still, the robustness and accuracy of feature-based methods are subject to the number of detected keypoints and the accuracy with which keypoint correspondences can be determined \citep{multiple_view_geometry}. Thus, feature-based methods may be less reliable when few correspondences exist due to large view differences or where the extracted features are not sufficiently salient due to image-specific lighting or noise. Methods that rely on estimating a differential homography between subsequent video frames and other features have been proposed \citep{road_plane, markerless}. Indeed, the authors note that their methods fail under some lighting conditions. Recent methods leverage deep neural networks to regress a parameterisation of the homography matrix directly \citep{unsupervised_deep_homog, deep_homo, deep_homo_dynamic, video_stab}, using features directly computed from a tuple of image patches for which a homography estimate is desired.
        
    \subsection{Sports field registration}
    \label{sec:sports_field_registration}
        Various recent methods of sports field registration identify corresponding features between the field template and camera image with the use of deep (convolutional) neural networks \citep{robust, kpsfr, grass_band, deep_structured, real_time_pose, TVcalib, synthetic, end_end}. These are subsequently used to estimate a mapping between the image and field template. This estimation may be achieved with the DLT \citep{multiple_view_geometry} or RANSAC \citep{ransac} algorithms \citep{robust, kpsfr, real_time_pose, TVcalib}. Some methods refine this estimate with, or otherwise rely entirely on, a combination of the following: regressing the homography directly from the input image and field model with a deep neural network \citep{self_supervised, optimising}, obtaining an estimate for the camera pose by matching with a feature-pose database \citep{end_end, synthetic}, iterative optimisation of the camera pose or homography based on re-projection error or some other metric \citep{end_end, TVcalib, optimising, synthetic, self_supervised}, or the use of a Markov Random Field (MRF) \citep{deep_structured}.

        Using a feature-pose database is cumbersome and will almost always require additional optimisation. Furthermore, methods that rely on optimisation tend to be slow, from experiments with \citep{TVcalib}. Similarly, the authors of \citep{deep_structured} report many average iterations and an average inference time of $0.44$ seconds with their MRF-based method when applied to soccer field registration.
        
        In comparison, keypoint-detection-based models are attractive due to their relatively small computational footprint and the potential to use the detected keypoints in a Bayesian framework, which is the approach in this paper.

    \subsection{Tracking}
    \label{sec:tracking_by_detection}        
        Few existing methods exploit the temporal consistency between subsequent video frames, with some exceptions being found in \citep{robust, real_time_pose, markerless, road_plane}. The differential homography between video frames is used in \citep{road_plane, markerless} for road plane detection (which is important in autonomous driving applications) with optical flow and human-assisted keypoint-less tracking, respectively. However, neither of these methods considers the differential homography in a Bayesian setting. \citep{robust} make use of online homography refinement by minimising two loss functions, one of which also takes into account the relative homography $\mathbf{\hat{H}}_t\mathbf{\hat{H}}^{-1}_{t-1}$ between the current and previous frame as in \citep{road_plane, markerless}. Once again, this is not performed with a Bayesian treatment of the homography. Finally, \citep{real_time_pose} makes use of player positions and field keypoints detected by a U-Net architecture \citep{unet}. A homography estimate is obtained for each frame in a sequence and decomposed to estimate camera intrinsic and extrinsic parameters. A condensation particle filter \citep{condensation} is applied, which only considers and enforces temporal consistency on the external camera parameters i.e. the dynamics model applied is $\mathbf{\left[R\mid C\right]}_t = \mathbf{\left[R\mid C\right]}_{t-1} + \mathcal{N}(\mathbf{0}, \mathbf{\Sigma})$ with particle weights obtained by a re-projection metric. While this method employs the Bayesian particle filter framework, the dynamics model is unsuitable. Particularly, camera movement is not effectively modelled: changes in the estimated pose are modelled as noise. Additionally, keypoint uncertainty is not taken into account. Indeed, a heuristic measure is required to determine when the filter should be re-initialised after it inevitably diverges.
        
        To the best of the authors' knowledge, the literature has not explored a Bayesian approach that explicitly incorporates homography, field template and keypoint measurement uncertainty while also modelling relative camera motion.
        
\section{Approach}
\label{sec:approach}
    The proposed approach, Bayesian Homography Inference from Tracked Keypoints (BHITK), is inspired by recent developments in tracking-by-detection methods. Specifically, those which employ a form of camera motion compensation. It has been shown that tracking performance can be improved by transforming bounding boxes forecasted at time $t-1$ such that they align more closely with detections at time $t$. This transformation effectively estimates and corrects for the non-stationarity in measurements induced by camera motion. This is performed in \citep{bells, invisible, modelling_ambiguous} with the image registration algorithm in \citep{parametric}, which estimates a non-linear mapping of pixels from one frame to the next. Another method makes use of ORB \citep{orb} and RANSAC \citep{ransac} to quickly align subsequent frames \citep{giao}. The global motion compensation technique of the Video Stabilisation module of OpenCV \citep{opencv} is used instead in \citep{botsort}. Its use is motivated by sparse optical flow features and translation-based local outlier rejection, which allows the resulting affine matrix estimated by RANSAC to be largely unaffected by dynamic objects. Since this method is more focused on background motion, it is ideal for use in the present case where several dynamic objects (e.g. soccer players, the ball, referees and spectator movement) are expected to be present.

    As noted by \citep{robust}, sparse features due to generally texture-less sports fields, narrow camera field of view, and occlusion by players represent the most significant challenges to keypoint-detection-based methods since these challenges lead to fewer detected keypoints and consequently a less robust homography estimate. This work proposes a dynamics model that explicitly relates image keypoint positions from one frame to the next. A relation between subsequent homographies is derived from this, appropriately considering camera motion. A Kalman filter framework with linear and non-linear components is used, encompassing the homography as part of the state vector. Thus, even when few or no keypoints are detected in narrow-field-of-view or occluded situations, the homography estimate forecasted by the dynamics model -- which is independent of specific field template keypoints -- may still be reasonably accurate due to the incorporation of the history of keypoints that were visible up until that time, and the possible estimation of out-of-frame keypoint positions. Furthermore, keypoint noise is modelled explicitly, contributing to even more accurate homography estimation. Contrary to the status quo, RANSAC \citep{ransac} is only used to obtain an initial homography estimate. Thereafter, the homography is inferred solely from the dynamics and measurement models and the fusion of the measured keypoint statistics. Whereas RANSAC considers keypoint measurements to be noisy point estimates, the proposed method instead considers the entire estimated distribution of each keypoint. The approach is flexible and can extend existing keypoint detection methods. Table \ref{tab:lit_compare} summarises the contributions of BHITK compared to related methods in the literature. \begin{table}[bht!]
            \caption{A comparison between the approaches of BHITK and related methods.}
            \newcolumntype{C}{>{\centering\arraybackslash} m{0.18\columnwidth} }
            \centering
            \begin{tabular}{p{0.2\columnwidth} m{0.25\columnwidth} C C}
            \\\toprule
                \centering Method & \centering Relative (inter-frame) homography or pose & Bayesian framework & Keypoint uncertainty (fully Bayesian approach)$^*$\\\toprule
                \cite{road_plane} & \centering\checkmark & \text{\sffamily X} & \text{\sffamily X} \\\midrule
                \cite{markerless} & \centering\checkmark & \text{\sffamily X} & \text{\sffamily X} \\\midrule
                \cite{robust} & \centering\checkmark & \text{\sffamily X} & \text{\sffamily X} \\\midrule
                \cite{real_time_pose} & \centering modelled inappropriately & \checkmark & \text{\sffamily X} \\\midrule
                BHITK (proposed) & \centering\checkmark & \checkmark & \checkmark \\ \bottomrule
                \multicolumn{4}{p{\columnwidth}}{$^*$\footnotesize Equivalently, whether RANSAC is used to obtain the homography from noisy point estimates (indicated with a cross) or whether the homography is inferred with the fusion of different keypoint distributions (indicated with a checkmark).}\\
                \end{tabular}
            \label{tab:lit_compare}
        \end{table}

    \subsection{Derivations}
    \label{sec:the_model}
        Given a set of $N$ known field template keypoints represented by normalised homogeneous world coordinates $\left\{\mathbf{X}^{F, j}\in\mathbb{P}^2|1 \leq j \leq N\right\}$, and a set of $N$ image keypoints represented by normalised homogeneous image coordinates $\left\{\mathbf{x}^{I, j}\in\mathbb{P}^2|1 \leq j \leq N\right\}$, the corresponding points in these sets are assumed to be coplanar in world coordinates. The goal is to estimate the homography $\mathbf{H}$, which relates keypoints in the image and their corresponding coordinates in the field template. Assume that an image motion $\mathbf{A}$ is available at each time step $t$ such that \begin{equation}
                \label{affine_trans}
                \mathbf{x}^{I, j}_t = \mathbf{A}_t\mathbf{x}^{I, j}_{t-1}.
            \end{equation}
        Furthermore, since field template keypoints are constant: \begin{equation}
            \label{const_keypoints}
                \mathbf{X}^{F,j}_t = \mathbf{X}^{F,j}_{t-1},
            \end{equation}
        where the dependence on time is retained since time-dependent random samples are later added to model the possible uncertainty of field template keypoint positions. From (\ref{homo_mapping}), the coordinates in $\mathbb{P}^2$ of an image keypoint $\mathbf{x}^{I, j}$ which corresponds to a field template keypoint $\mathbf{X}^{F, j}$ may be obtained by \begin{equation}
            \label{measurement_eq}
                \mathbf{x}^{I,j}_t = \mathbf{H}_t\mathbf{X}^{F,j}_t.
            \end{equation}
        Substituting (\ref{measurement_eq}) into (\ref{affine_trans}): \begin{equation*}
                \mathbf{x}^{I, j}_t = \mathbf{A}_t\mathbf{H}_{t-1}\mathbf{X}^{F, j}_{t-1}.
            \end{equation*}
        Making use of the relation in (\ref{const_keypoints}): \begin{equation*}
                \mathbf{x}^{I, j}_t = \mathbf{A}_t\mathbf{H}_{t-1}\mathbf{X}^{F, j}_{t}.
            \end{equation*}
        Finally, comparing this result with (\ref{measurement_eq}): \begin{equation}
                \label{homo_process}
                \mathbf{H}_t = \mathbf{A}_t\mathbf{H}_{t-1}.
            \end{equation}
            
        In practice, it is necessary to obtain the image coordinates in $\mathbb{R}^2$ represented by $\mathbf{x}^{I, j}_t\in \mathbb{P}^2$. This is achieved by normalising the homogeneous $\mathbf{x}^{I, j}_t$ with respect to its last element, as illustrated in (\ref{normalisation}). Let $\operatorname{norm}(\cdot)$ denote this normalisation, such that $\operatorname{norm}(\begin{bmatrix}
            x & y & z 
        \end{bmatrix}^\top) = \begin{bmatrix}
            x/z & y/z & 1
        \end{bmatrix}^\top$. Thus, (\ref{measurement_eq}) becomes \begin{equation}
            \label{normalised_measurement_eq}
                \mathbf{x}^{I,j}_t = \operatorname{norm}\left(\mathbf{H}_t\mathbf{X}^{F,j}_t\right).
            \end{equation}
            
        $\mathbf{A}_t$ is an affine transformation matrix which allows for translation, rotation and uniform scaling in the x and y image dimensions. It is estimated at each time step with the global motion compensation method of OpenCV \citep{opencv}: \begin{equation*}
                \mathbf{\hat{A}}_t = \left[\begin{array}{c|c} 
                                \mathbf{A}_t^u\in \mathbb{R}^{2\times 2} & \mathbf{b}_t\in \mathbb{R}^{2\times 1} \\
                                \mathbf{0}^{1\times 2} & 1
                                \end{array}\right],
            \end{equation*}
        where $\mathbf{A}_t^u$ is a rotation and scaling matrix and $\mathbf{b}_t$ is a translation vector, such that, for arbitrary vectors, $\mathbf{x}$ and $\mathbf{y}$ in $\mathbb{R}^2$, \begin{equation*}
                \mathbf{y}=\mathbf{A}_t^u \mathbf{x}+\mathbf{b}_t
            \end{equation*}
        is equivalent to \begin{equation*}
                \begin{bmatrix}
                    \mathbf{y} \\ 1
                \end{bmatrix} = \mathbf{\hat{A}}_t\begin{bmatrix}
                    \mathbf{x} \\ 1
                \end{bmatrix}
            \end{equation*} in homogeneous coordinates. It is easily shown that, for an arbitrary $\mathbf{x}\in\mathbb{P}^2$, $\operatorname{norm}(\mathbf{A}_t\mathbf{x})=\mathbf{A}_t\operatorname{norm}(\mathbf{x})$. Therefore, the relation obtained in (\ref{homo_process}) remains valid despite the alteration to (\ref{measurement_eq}) given in (\ref{normalised_measurement_eq}). Note that this is not generally the case if $\mathbf{A}_t$ is replaced with, e.g. an inter-frame homography (i.e. a homography that provides a mapping between the current and previous frames) where the last row is not $\begin{bmatrix}
            0 & 0 & 1
        \end{bmatrix}$, which would require additional normalisation. Therefore, the proposed affine transformation enables the relationship between subsequent homographies to be expressed and precludes normalisation in (\ref{affine_trans}), maintaining linearity in the keypoint dynamics model. Another advantage of using the proposed affine transformation is that it is independent of detecting specific, pre-defined keypoints. Thus, the absence of any number of such keypoints is assumed not to affect the dynamics model significantly. Nevertheless, incorporating a robust, global transformation of keypoints between subsequent frames (which may also be non-linear) into the state vector is a compelling prospect but deemed a topic for future research. For now, it is assumed that uncertainties in the estimation of $\mathbf{A}_t$ are largely mitigated by modelling (\ref{affine_trans}) and (\ref{homo_process}) as stochastic processes. Uncertainty in field template keypoint positions may also be specified, although this inclusion may only benefit practical applications where field dimensions vary. The dynamics model is thus obtained by treating (\ref{affine_trans}), (\ref{const_keypoints}) and (\ref{homo_process}) as random variables: \begin{gather}
                \mathbf{x}^{I, j}_t = \mathbf{A}_t\mathbf{x}^{I, j}_{t-1} + \mathbf{w}^{I,j}_t \label{kp_dynamics},\\
                \mathbf{X}^{F,j}_t = \mathbf{X}^{F,j}_{t-1} + \mathbf{w}^{F,j}_t \label{fieldpt_dynamics},\\
                \mathbf{H}_t = \mathbf{A}_t\mathbf{H}_{t-1} + \mathbf{W}^H_t, \label{homog_dynamics}
            \end{gather}
        where $\mathbf{w}^{I, j}_t \sim \mathcal{N}\left(\mathbf{0}, \mathbf{\Sigma}^{I, j}\right)$ and $\mathbf{w}^{F, j}_t \sim \mathcal{N}\left(\mathbf{0}, \mathbf{\Sigma}^{F, j}\right)$. The elements of $\mathbf{W}^H_t$ are drawn from $\mathcal{N}\left(\mathbf{0}, \mathbf{\Sigma}^H\in\mathbb{R}^{9\times 9}\right)$. Similarly, the measurement model is obtained from (\ref{normalised_measurement_eq}):
            \begin{equation}
                \label{measurement_model}
                \mathbf{x}^{I,j}_t = \operatorname{norm}\left(\mathbf{H}_t\mathbf{X}^{F,j}_t\right) + \mathbf{w}^{M, j}_t,
            \end{equation}
        where $\mathbf{w}^{M, j}_t \sim \mathcal{N}(\mathbf{0}, \mathbf{\Sigma}^{M, j})$. The last element of each homogeneous coordinate $\mathbf{x}^{I, j}_t$ and $\mathbf{X}^{F,j}_t$ is always 1. For the remainder of this paper, these coordinates are assumed to be transformed to $\mathbb{R}^2$ by simply omitting their last elements. Therefore, the distributions from which $\mathbf{w}^{I,j}_t$, $\mathbf{w}^{F,j}_t$ and $\mathbf{w}^{M, j}_t$ are drawn are also considered elements of $\mathbb{R}^2$, with corresponding covariance matrices in $\mathbb{R}^{2\times 2}$.

\begin{figure}[hbt!]
            \centering
            \begin{tikzpicture}[
    squarednode/.style={rectangle, draw=black, thick, minimum size=5mm},
    arrow/.style={-{stealth[scale=1.2]}, thick},
    ]
    \node[squarednode] (lkf) {Keypoint LKF};
    \node (measurements) [left=0.4cm of lkf] {$\mathbf{y}^{I}_t$};
    \node (affine) [above=0.5cm of measurements] {$\mathbf{\hat{A}}_t$};
    \node[squarednode] (ekf) [right=0.5cm of lkf] {Homography EKF};
    \node (homography) [right=0.4cm of ekf] {$\mathbf{\hat{H}}_t$};
    \node (keypoints) [below=0.5cm of homography] {$\mathbf{\hat{x}}^{I}_t$};
    
    \draw[arrow] (affine) -| (lkf);
    \draw[arrow] (measurements) -- (lkf);
    \draw[arrow] (lkf) -- coordinate[midway](lekf) (ekf);
    \draw[arrow] (ekf) -- (homography);
    \draw[arrow] (lekf) |- (keypoints);
    \draw[arrow] (affine) -| (ekf);
\end{tikzpicture}
            \caption{Implemented Kalman filter framework. The first stage filters measured keypoint positions $\mathbf{y}^{I}_t$ according to the estimated affine transformation $\mathbf{\hat{A}}_t$. The EKF makes use of the filtered positions, $\mathbf{\hat{x}}^{I}_t$, and the estimated affine transformation to infer an estimate of the homography, $\mathbf{\hat{H}}_t$.}
            \label{fig:kalman_framework}
        \end{figure}
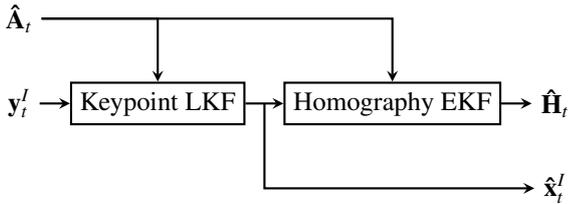

        The above dynamics and measurement models are used in a two-stage Kalman filer. Fig. \ref{fig:kalman_framework} concisely illustrates the roles of the estimated affine transformation $\mathbf{\hat{A}}_t$, keypoint measurements $\mathbf{y}^I_t$ and two-stage filter in the proposed approach. The filter stages are subsequently described in detail. 

    \subsection{Linear keypoint filter}
    \label{sec:lkf}
        The left-hand side of (\ref{kp_dynamics})--(\ref{homog_dynamics}) represents the state space under consideration. The state elements are not independent. A given set of keypoint measurements may be related to the state in two ways: either directly since the keypoint positions are part of the state vector or through the re-projection of the field keypoints to image keypoints by the homography. Although not strictly required for homography inference, image keypoint positions are retained as part of the state vector. This can improve the homography estimation since the filtered keypoint positions are likely more accurate, as long as the zero-mean Gaussian measurement noise assumption is reasonable and the process and measurement covariances are appropriately tuned. Hence, a two-stage approach is proposed. The first stage consists of a linear Kalman filter (LKF), which considers the image keypoints the only part of its state vector. The state vector takes the form $\mathbf{x}^I_t=\begin{bmatrix}
           \left(\mathbf{x}^{I,1}\right)^\top \cdots \left(\mathbf{x}^{I,j}\right)^\top
        \end{bmatrix}^\top_t$, where $\mathbf{x}^{I, j}\in\mathbb{R}^2, j \leq N$. Its dynamics are governed by (\ref{kp_dynamics}). The prediction step is implemented as follows: \begin{gather*}
                \hat{\mathbf{x}}^I_{t \mid t-1}=\tilde{\mathbf{A}}_t \hat{\mathbf{x}}^I_{t-1 \mid t-1}+\tilde{\mathbf{B}}_t, \\
                \mathbf{P}^I_{t \mid t-1}=\tilde{\mathbf{A}}_t \mathbf{P}^I_{t-1 \mid t-1} \tilde{\mathbf{A}}_t^{\top} + \mathbf{Q}^I,
            \end{gather*}
        where $\hat{\mathbf{x}}^I_{t \mid t-1}$ and $\hat{\mathbf{x}}^I_{t-1 \mid t-1}$ denote the estimates of the means of the predicted and filtered states at time $t-1$, respectively. Similarly, $\mathbf{P}^I_{t \mid t-1}$ and $\mathbf{P}^I_{t-1 \mid t-1}$ denote the predicted and filtered estimates of the state covariance matrix. $\mathbf{Q}^I$ represents the process noise covariance matrix. Finally, similar to \citep{botsort}: \begin{equation*}
                \tilde{\mathbf{A}}_t = \begin{bmatrix}
                    \mathbf{A}_t^u & \mathbf{0} & \mathbf{0}\\
                    \mathbf{0} & \ddots & \mathbf{0}\\
                    \mathbf{0} & \mathbf{0} & \mathbf{A}_t^u
                \end{bmatrix}, 
                \tilde{\mathbf{B}}_t = \begin{bmatrix}
                    \mathbf{b}_t \\
                    \vdots \\
                    \mathbf{b}_t
                \end{bmatrix}.
            \end{equation*} 

        Upon receiving $K$ keypoint measurements $\mathbf{y}_t^I=\begin{bmatrix}
            \left(\mathbf{y}^{I,j}_1\right)^\top \cdots \left(\mathbf{y}^{I,j}_k\right)^\top
        \end{bmatrix}^\top_t$, where $\mathbf{y}^{I,j}_k\in\mathbb{R}^2, k\leq K$, the update step is performed: \begin{gather*}
                 \mathbf{K}_t=\mathbf{P}^I_{t \mid t-1}\mathbf{H}_t^{\top}\left(\mathbf{H}_t \mathbf{P}^I_{t \mid t-1}\mathbf{H}_t^{\top}+\mathbf{R}^I\right)^{-1}, \\
                 \hat{\mathbf{x}}_{t \mid t}^I=\hat{\mathbf{x}}_{t \mid t-1}^I+\mathbf{K}_t\left(\mathbf{y}_t^I-\mathbf{H}_t \hat{\mathbf{x}}_{t \mid t-1}^I\right), \\
                 \mathbf{P}^I_{t \mid t}=\left(\mathbf{I}-\mathbf{K}_t \mathbf{H}_t\right) \mathbf{P}^I_{t \mid t-1},
            \end{gather*} where $\mathbf{R}^I$ is the measurement noise covariance matrix and $\mathbf{H}_t\in\mathbb{R}^{2K\times 2N}$ is a matrix which consists of sub-matrices $\mathbf{h}_{k, j}\in\mathbb{R}^{2\times 2}$ that relate the state to the measurements:
            \begin{equation*}
                \mathbf{h}_{k, j} = \begin{cases}
                    \mathbf{I}, \text{ if } \exists\mathbf{y}^{I,j}_k\in \mathbf{y}_t^I,\\
                    \mathbf{0}, \text{ otherwise},
                \end{cases}
            \end{equation*}
            where $\mathbf{I}$ is the identity matrix.

    \subsection{Non-linear homography filter}
    \label{sec:homog_filter}
        The second stage of the proposed method directly incorporates the homography, in addition to the field template keypoints, into its state vector $\mathbf{x}^{FH}_t=\begin{bmatrix}
           \left(\mathbf{X}^{F,1}\right)^\top \cdots \left(\mathbf{X}^{F,j}\right)^\top
           & \mathbf{h}_1^\top & \mathbf{h}_2^\top & \mathbf{h}_3^\top
            \end{bmatrix}^\top_t$, where $\mathbf{X}^{F, j}\in\mathbb{R}^2, j \leq N$ and $\mathbf{h}_1$, $\mathbf{h}_2$ and $\mathbf{h}_3$ respectively denote the first, second and third columns of the homography matrix. Its dynamics are governed by (\ref{fieldpt_dynamics}) and (\ref{homog_dynamics}), which are linear processes. The prediction step is thus performed by
            \begin{gather*}
                \hat{\mathbf{x}}^{FH}_{t \mid t-1}=\tilde{\mathbf{M}}_t \hat{\mathbf{x}}^{FH}_{t-1 \mid t-1},\\
                \mathbf{P}^{FH}_{t \mid t-1}=\tilde{\mathbf{M}}_t \mathbf{P}^{FH}_{t-1 \mid t-1} \tilde{\mathbf{M}}_t^{\top} + \mathbf{Q}^{FH},
            \end{gather*} where $\mathbf{P}^{FH}$ and $\mathbf{Q}^{FH}$ denote the state and process covariance matrices, respectively. Furthermore,
            \begin{equation*}
                \tilde{\mathbf{M}}_t = \begin{bmatrix}
                    \mathbf{I}^{2N\times 2N} & \mathbf{0}\\
                    \mathbf{0} & \mathbf{\hat{A}}_t
                \end{bmatrix}.
            \end{equation*}
            
        The non-linear measurement model (\ref{measurement_model}) requires the transformation of the state by (\ref{normalised_measurement_eq}). This may be performed whilst retaining relatively high-order moments using the Unscented Transform in the Unscented Kalman Filter (UKF) \citep{ukf}. However, since the relative homography between frames is expected to be small, the Extended Kalman Filter (EKF) approach is used instead by linearising around the current state estimate. Enforcing this belief in the UKF requires tuning its hyper-parameters (i.e. how close to the mean sigma points are sampled), which is avoided. Using the UKF slightly degraded performance due to significant errors before convergence. The update step is therefore performed as follows: \begin{gather*}
                 \mathbf{K}^{FH}_t=\mathbf{P}^{FH}_{t \mid t-1}\mathbf{J}_t^{\top}\left(\mathbf{J}_t \mathbf{P}^{FH}_{t \mid t-1}\mathbf{J}_t^{\top}+\mathbf{P}^I_{t\mid t}\right)^{-1}, \\
                 \hat{\mathbf{x}}^{FH}_{t \mid t}=\hat{\mathbf{x}}^{FH}_{t \mid t-1}+\mathbf{K}^{FH}_t\left(\hat{\mathbf{x}}_{t \mid t}^I-\operatorname{h}\left(\hat{\mathbf{x}}_{t \mid t-1}^{FH}\right)\right), \\
                 \mathbf{P}^{FH}_{t \mid t}=\left(\mathbf{I}-\mathbf{K}^{FH}_t \mathbf{J}_t\right) \mathbf{P}^{FH}_{t \mid t-1},
            \end{gather*} where $\operatorname{h}(\cdot)$ represents (\ref{normalised_measurement_eq}), $\hat{\mathbf{x}}_{t \mid t-1}^{FH}$ is augmented with a one in (\ref{normalised_measurement_eq}) to transform it to $\mathbb{P}^2$, and $\mathbf{J}_t$ is the Jacobian of $\operatorname{h}\left(\hat{\mathbf{x}}_{t \mid t-1}^{FH}\right)$ with respect to each of the state elements.

        The proposed approach is adaptable to any keypoint detection method. Furthermore, the state and measurement models may be expanded to incorporate image distortion parameters. However, modelling distortion with a single-parameter division model as in \citep{distortion} slightly degraded performance and is therefore not considered, although such parameters may be useful in some practical applications.
        
\section{Experiments}

The following section describes the experiments used to evaluate BHITK in detail. This includes details on the practical implementation, parameter settings, datasets and evaluation metrics.

\label{sec:Experiments}
    \subsection{Practical considerations}
    \label{sec:practical_considerations}
        The state mean estimates $\hat{\mathbf{x}}^I_{0 \mid 0}$ and $\hat{\mathbf{x}}^{FH}_{0 \mid 0}$ are initialised with the first measured keypoint positions, known field template positions and the initial homography estimate obtained with RANSAC. Since the homography matrix is determined up to a scale (\ref{homo_mapping}), the initial estimate is normalised with respect to $h_{33}$. The Kalman filter state vector excludes this element ($h_{33}$). 
        
        For the current purposes, $\mathbf{\Sigma}^{F, j}=\mathbf{0}\forall j$. The other process covariance matrices $\mathbf{\Sigma}^{I, j}$ and $\mathbf{\Sigma}^{H}$ are estimated empirically from the training data using the estimated affine transformation between training video frames, the ground truth keypoint positions and the ground truth homography annotations. Specifically, the mean-squared error (MSE) is used to estimate the variance of $\mathbf{\Sigma}^{I, j}$ and $\mathbf{\Sigma}^{H}$ in each state dimension, and the mean of the product of the errors of different state dimensions are used to estimate the covariances. 
        
        The measurement covariance matrix $\mathbf{\Sigma}^{M, j}$ of each keypoint $j$ is estimated similarly using the measured and ground truth keypoint positions from the training set. The measured keypoint positions are used with RANSAC to produce a homography estimate for each training frame. These estimates are used with the ground truth homography annotations to estimate the covariance matrix with which the initial homography estimate is initialised. These are the only covariance matrices dependent on the keypoint detection method. 
        
        The process covariance matrices $\mathbf{Q}^I$ and $\mathbf{Q}^{FH}$, and the measurement covariance matrix $\mathbf{R}^I$ are constructed by concatenating the applicable covariance matrices obtained in the manner described above diagonally. The estimation of the covariance between distinct keypoints is complicated by the fact that keypoints do not always co-occur in the same image. Thus, independence between distinct keypoints is assumed. Finally, independence between distinct field template keypoints and between field template keypoints and the homography is also assumed. This follows from treating the field template positions as known ($\mathbf{\Sigma}^{F, j}=\mathbf{0}\forall j$), which is justified in the present case since the ground truth homography annotations are also obtained with this assumption.

        While the discussion of the LKF and EKF in \ref{sec:lkf} and \ref{sec:homog_filter} imply that all of the known field template keypoint positions have corresponding image keypoints in $\mathbf{x}^I_t$, only the keypoints which have been measured at times prior to and including the current time step $t$ are used in the EKF update step. Furthermore, the best results have been obtained when only the keypoints measured at the current time step $t$ are used in the EKF update step. However, in the case of sparse keypoint positions, it may be helpful to initialise all image keypoint positions through the initial homography estimate and use all of the estimated keypoint positions in the EKF update, especially since the covariance estimates of those keypoints that have not been measured recently would be larger than that of those that have. The EKF takes this uncertainty into account.

        \subsubsection{Assumptions}
            The following summarises the major assumptions of the proposed BHITK method:

            \begin{itemize}
                \item The true playing field dimensions correspond to the template field dimensions.
                \item Equations (\ref{kp_dynamics})--(\ref{homog_dynamics}) are reasonable approximations of the true state dynamics i.e. nonlinearities in the true state dynamics are negligible.
                \item Equation \ref{measurement_model} is a reasonable approximation of the true observation model i.e. lens distortion effects are negligible.
                \item The process and measurements noise terms in (\ref{kp_dynamics})--(\ref{measurement_model}) are reasonably approximated by zero-mean Gaussian distributions.
                \item The linearisation error of the EKF is smaller than the measurement noise.
                \item Distinct keypoints are independent of one another in the image and ground plane, and the homography is independent of specific field template positions (see the discussion in \ref{sec:practical_considerations}).
                
            \end{itemize}
    
    \subsection{Datasets}
    \label{sec:datasets}
        \subsubsection{WC14 dataset} The WorldCup (WC14) dataset is typically used to evaluate soccer field registration \citep{kpsfr, robust, TVcalib, deep_structured, real_time_pose, optimising, synthetic, end_end, self_supervised}. It consists of 209 image-homography pairs in a training set and 186 in a test set. The images were obtained from broadcast television videos of the 2014 FIFA World Cup. The ground truth homography matrices are labelled manually. Unfortunately, as already noted \citep{TVcalib}, the annotation of homography matrices is biased since the entire field is usually not visible in any given image. This problem is exacerbated by using too few ground-truth keypoints, i.e. a sparse keypoint annotation template focusing only on certain parts of the field. This is illustrated in Fig. \ref{fig:wc14_example}, which shows an example of low-quality homography annotation in the WC14 dataset by re-projecting the grass band keypoints proposed in \citep{grass_band} with the annotated homography. Notice that the keypoints do not align well with the grass bands, particularly those further away from the penalty area. Inadequate homography annotations undermine the reliability of the Intersection over Union (IoU) metrics often used to evaluate soccer field registration methods \citep{TVcalib, real_time_pose}.

        \subsubsection{TS-WorldCup dataset} The TS-WorldCup dataset (TSWC) was introduced to augment the WC14 dataset since the WC14 dataset is relatively small \citep{kpsfr}. Unlike the WC14 dataset, the TSWC dataset consists of consecutive frames from 43 2014 and 2018 Soccer World Cup event videos. It contains 2925 and 887 images in its training and test sets. Similar to the WC14 dataset, the TSWC dataset suffers from annotation bias, albeit somewhat. This is illustrated in Fig. \ref{fig:ts_example}, where the annotation error is especially visible towards the bottom of the image.

        \subsubsection{CARWC dataset}\label{carwc} 
\begin{figure*}[hbt!]
                \centering
                \caption{Re-projected keypoints using WC14 and custom (CARWC) annotated homographies. The re-projection error of the WC14 annotation is most noticeable when considering the alignment with the right-most grass band.}
                    \begin{subfigure}[]{\textwidth}
                        \centering
                        \includegraphics[width=\textwidth]{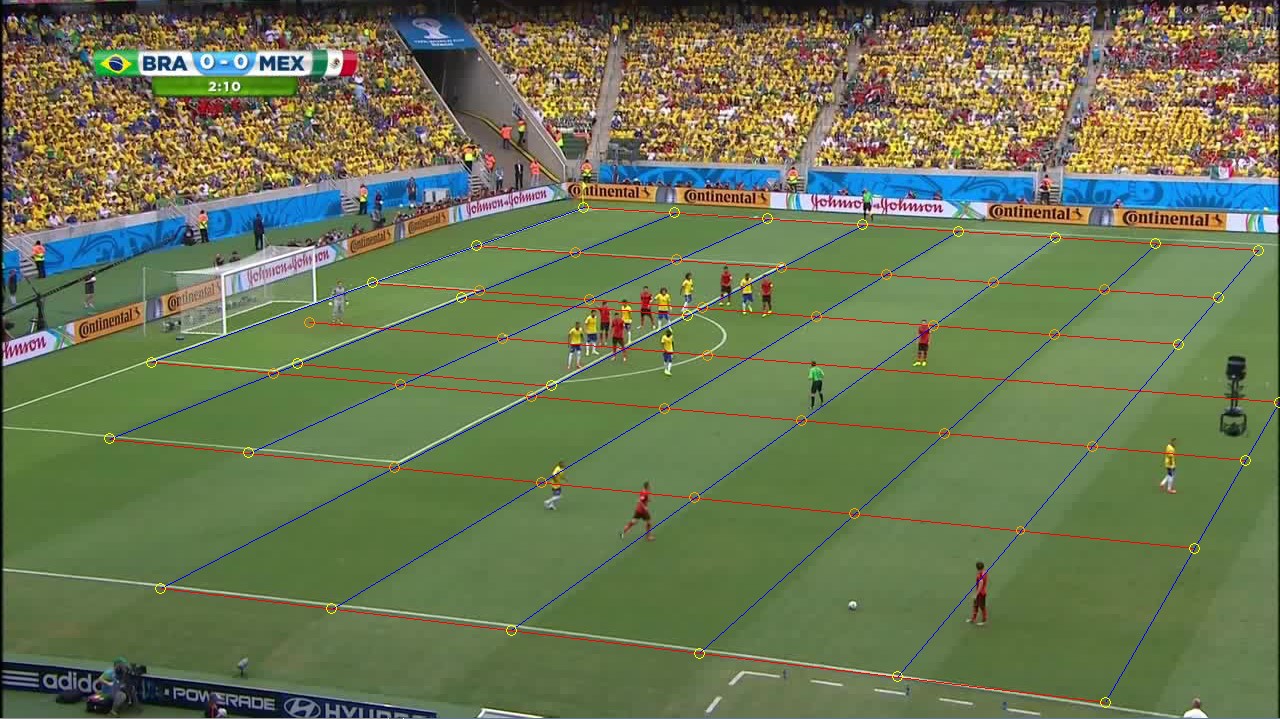}
                        \caption{WC14.}
                        \label{fig:wc14_example}
                    \end{subfigure}
                    \hfill
                    \begin{subfigure}[]{\textwidth}
                        \centering
                        \includegraphics[width=\textwidth]{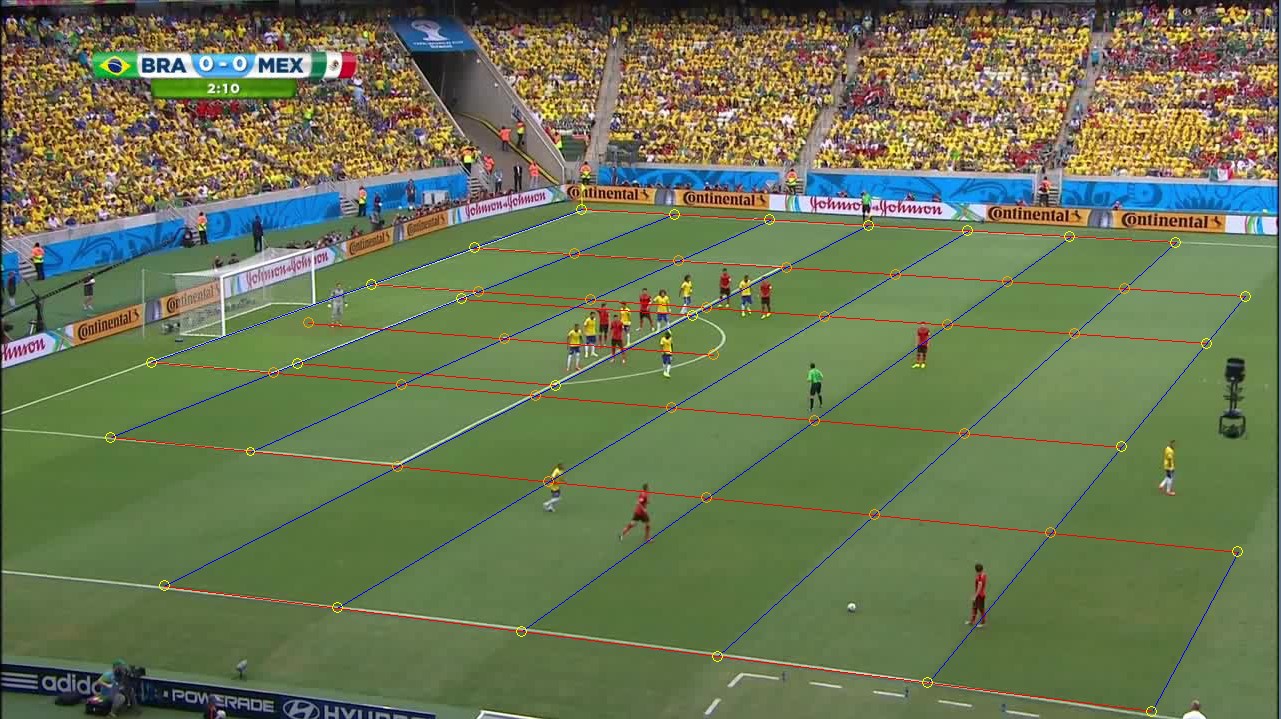}
                        \caption{CARWC.}
                        \label{fig:custom_wc14_example}
                    \end{subfigure}
                \label{fig:wc14_comparison}
            \end{figure*}

\begin{figure*}[hbt!]
                \centering
                \caption{Re-projected keypoints using TSWC and custom (CARWC) annotated homographies. The re-projection error of the TSWC annotation is most noticeable when considering the alignment with the bottom length-wise horizontal field line.}
                    \begin{subfigure}[]{\textwidth}
                        \centering
                        \includegraphics[width=\textwidth]{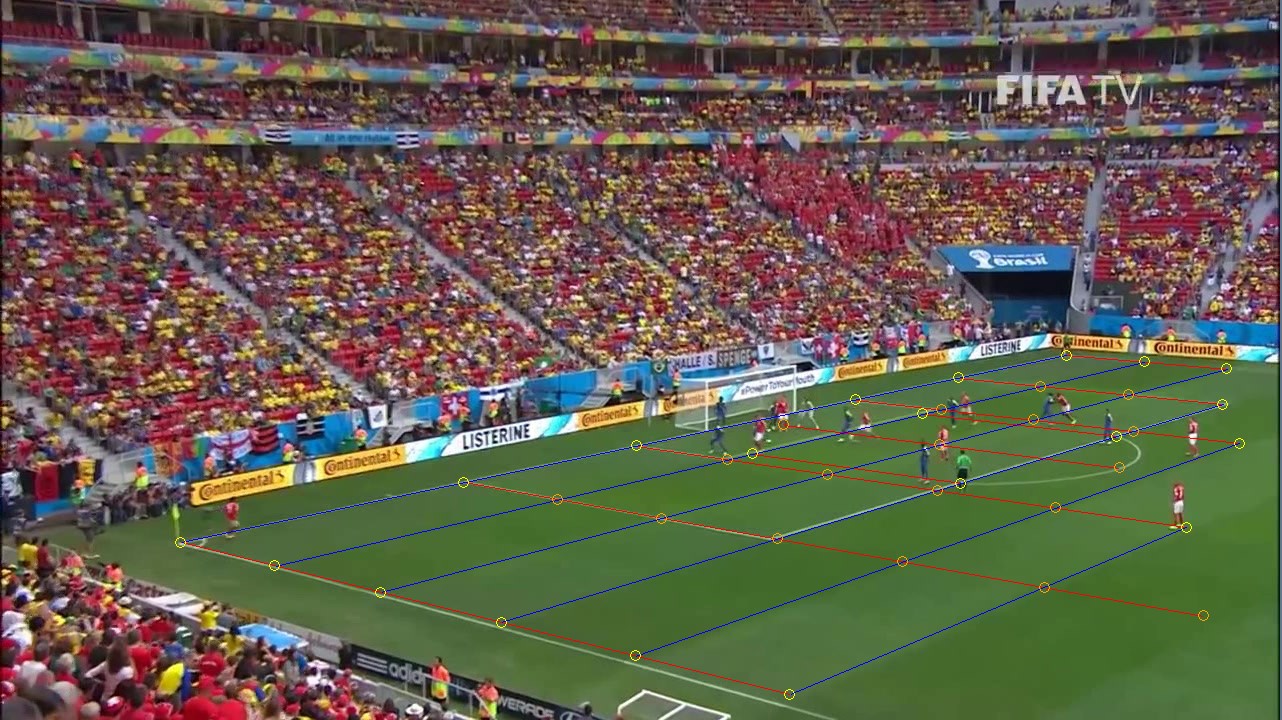}
                        \caption{TSWC.}
                        \label{fig:ts_example}
                    \end{subfigure}
                    \hfill
                    \begin{subfigure}[]{\textwidth}
                        \centering
                        \includegraphics[width=\textwidth]{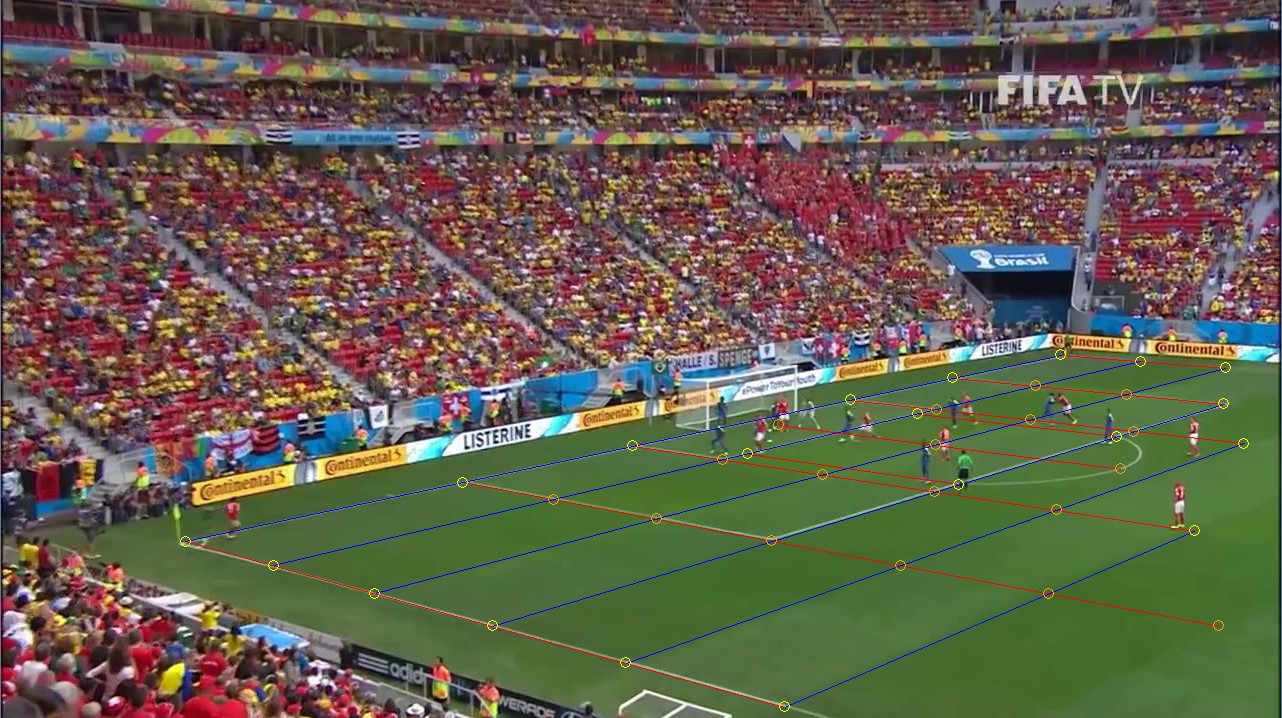}
                        \caption{CARWC.}
                        \label{fig:custom_ts_example}
                    \end{subfigure}
                \label{fig:ts_comparison}
            \end{figure*}

In this work, a consolidated and refined WorldCup (CARWC) dataset is introduced. The dataset combines the WC14 and TSWC datasets. Additionally, all images are re-annotated with the help of publicly available custom homography refinement software developed by the principal author of this paper, which makes use of image deformation methods \citep{deformation} to ease the annotation process and is released along with the CARWC dataset\footnote{\url{https://github.com/Paulkie99/KeypointAnnotator}.}. The grass band keypoints proposed in \citep{grass_band} are used during annotation. These keypoints are selected since they are dense while also retaining semantic meaning. There is a total of 147 grass band keypoints across the field. In comparison, \citep{robust} uses only 91 keypoints spread uniformly across the field. The grass band keypoints are placed in semantically meaningful locations, which could aid in their detection. While uniform keypoints do not necessarily occur at the intersections of lines or other distinguishable field markings, the grass band keypoints mainly occur at the intersection of grass bands with some other field marking (which may be extended, e.g. the horizontal lines of the penalty box, with some exceptions). This also makes these keypoints easier to identify during annotation. Fig. \ref{fig:custom_wc14_example} and Fig. \ref{fig:custom_ts_example} illustrate two examples of the refined annotations.

        The ground truth keypoint position labels are also included along with the homography annotations. This could allow for the investigation of image distortion in future work.

        To obtain a sense of scale for the process covariance matrices of the CARWC training set, the mean matrix entries over all $j$ of $\mathbf{\Sigma}^{I, j}$, estimated as described in \ref{sec:practical_considerations}, are \begin{equation*}
                \begin{bmatrix}
                     4.95 & -0.06\\
                    -0.06 & 0.95
                \end{bmatrix}.
            \end{equation*}
        Similarly, for $\mathbf{\Sigma}^{H}$, the mean entries are \begin{equation*}
        \begingroup
        \footnotesize
        \setlength\arraycolsep{4pt}
        \begin{bmatrix}
                 3.35 & 0.05 & 0.00 & 1.58 & 0.27 & 0.00 & -373 &  12.77\\
                 0.05 & 0.01 & 0.00 & 0.02 & 0.00 & 0.00 & -4.66 & -0.08\\
                 0.00 & 0.00 & 0.00 & 0.00 & 0.00 & 0.00 & -0.10 & 0.00\\
                 1.58 & 0.02 & 0.00 & 0.77 & 0.13 & 0.00 & -179 & 6.45 \\ 
                 0.27 & 0.00 & 0.00 & 0.13 &  0.03 & 0.00 & -30.11 &  0.94  \\ 
                 0.00 & 0.00 & 0.00 & 0.00 & 0.00 & 0.00 & 0.05 & 0.00 \\ 
                -373 & -4.66 & -0.10 & -179 & -30.11 & 0.05 & 41943 & -1458 \\ 
                 12.77 & -0.08 & 0.00 & 6.45 & 0.94 & 0.00 & -1458 & 81.31 \\ 
        \end{bmatrix},
        \endgroup
    \end{equation*} where covariances associated with $h_{33}$ have been omitted because this element is not included in the Kalman filter state vector as discussed in \ref{sec:practical_considerations}.
        
    \subsection{Baselines}
    \label{sec:baselines}
    The proposed method will be evaluated primarily by comparison with two state-of-the-art keypoint-detection-based methods, namely those of \citep{robust} and \citep{kpsfr}. 

\begin{figure}[hbt!]
            \centering
            \includegraphics[width=\columnwidth]{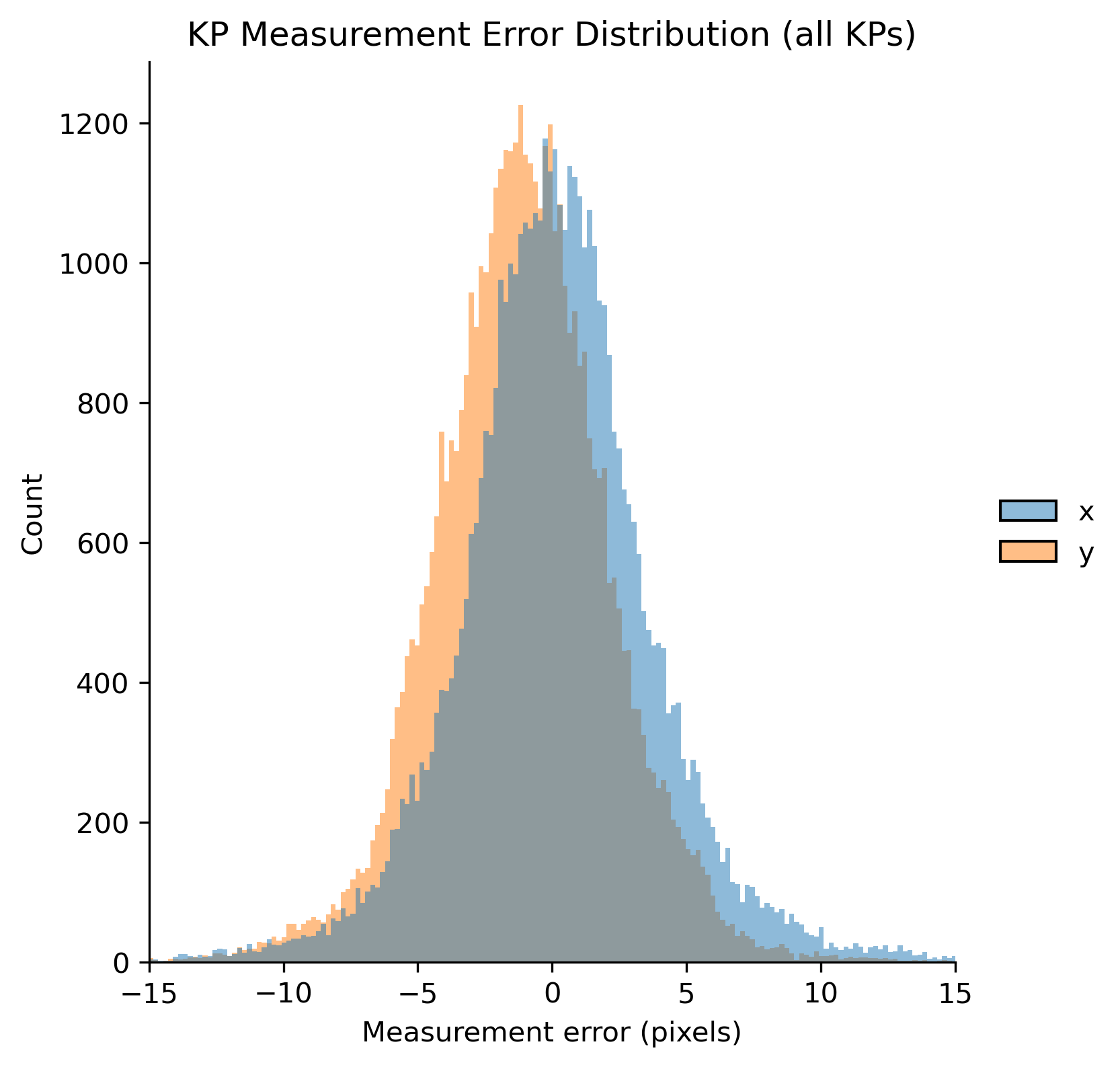}
            \caption{Keypoint (KP) measurement error distribution in the x- and y dimensions with the baseline model of \citep{robust}, trained and evaluated on the CARWC training set.}
            \label{fig:kp_meas_dist}
        \end{figure}

    The method proposed by \citep{robust} makes use of a ResNet-18-based \citep{resnet}, U-Net-like \citep{unet} architecture with non-local blocks \citep{nonlocal} and dilated convolutions. They propose the use of 91 keypoints spread uniformly across the field. In a multi-task learning approach, their method simultaneously predicts keypoints and dense features defined as the normalised distance of non-line or non-region pixels to the nearest line or region pixel in the image (referred to as line and region features, respectively). Finally, they use an online refinement scheme that considers these predicted features and the homography estimates for the current and previous frames. The re-implementation of their method in \citep{kpsfr} is used, which does not include dense feature regression or online refinement (i.e. performs keypoint detection only), and expanded with dense feature regression and online refinement, where the hyper-parameters for online refinement are set as in \citep{robust}. Fig. \ref{fig:kp_meas_dist} shows the keypoint error distribution of the trained model on the CARWC training set -- it is clear that the zero-mean Gaussian assumption is reasonable. The median matrix entries over all $j$ of $\mathbf{\Sigma}^{M, j}$, estimated as described in \ref{sec:practical_considerations} using the method of \citep{robust} on the CARWC training set, are \begin{equation*}
            \begin{bmatrix}
                20.81 & -0.01\\
                -0.01 & 14.56
            \end{bmatrix}.
        \end{equation*} Similarly, the mean entries of the estimated covariance matrix of the homography obtained with RANSAC (used to initialise the homography elements of the Kalman filter state vector) are \begin{equation*}
        \begingroup
        \footnotesize
        \setlength\arraycolsep{1.8pt}
            \begin{bmatrix}
                254 &       1.76 &       0.06 &     123 &      22.81 &
             -0.03 &  -28795 &     947\\
       1.76 &       0.25 &       0.00   &       0.54 &       0.05 &
             0.00   &    -163 &      -1.17\\
       0.06 &       0.00   &       0.00   &       0.03 &       0.01 &
             0.00   &      -6.51 &       0.21\\
       123 &       0.54 &       0.03 &      60.23 &      11.22 &
             -0.02 &  -14023 &     472\\
       22.81 &       0.05 &       0.01 &      11.22 &       2.18 &
             0.00   &   -2605 &      87.42\\
       -0.03 &      0.00   &      0.00   &      -0.02 &      0.00   &
              0.00   &       3.85 &      -0.12\\
       -28795 &    -163 &      -6.51 &  -14023 &   -2605 &
              3.85 & 3280363 & -108856\\
       947 &      -1.17 &       0.21 &     472 &      87.42 &
             -0.12 & -108856  &    4186\\
            \end{bmatrix}, 
        \endgroup
        \end{equation*} where the elements associated with $h_{33}$ have been omitted as in \ref{carwc}.
    
    The method proposed by \citep{kpsfr}, referred to as KpSFR, uses a ResNet-34-based encoder-decoder architecture incorporating skip connections. Using the same keypoint template proposed by \citep{robust}, dynamic filter learning is used to predict keypoints. While state-of-the-art results are reported, the method requires pre-processed results, e.g. acquired by the method of \citep{robust}, to obtain keypoint identity encodings during inference. Ignoring this pre-requisite, the method uses approximately 73 million parameters, and inference occurs at approximately 1.5 frames per second\footnote{As measured on an NVIDIA RTX 3090 GPU with an AMD Ryzen 9 5900X CPU, with the following inference script: \url{https://github.com/ericsujw/KpSFR/blob/main/inference.py}.}. In comparison, the re-implementation of \citep{robust} requires approximately 42 million parameters and executes at approximately 50 frames per second \citep{robust}.

    Both of these networks are trained in a manner similar to \citep{robust}: with the Adam \citep{adam} optimiser ($\beta_1=0.9, \beta_2=0.999$) for 300 epochs, where the initial learning rate of $1e-4$ decays to $1e-5$ after 200 epochs. Training takes place on the CARWC training set.

    The performance obtained before and after augmentation with our proposed method is of particular interest. Variations of the method of \citep{robust} (brought about by changes to the training program, the number or distribution of keypoints in the field template) are augmented with our BHITK approach. While some of these variations improve over other baseline methods without BHITK, additional improvements are attained using BHITK. Thus, the Bayesian modelling approach enables performance improvements which may not be attainable through other means.

    It has been shown that using stochastic gradient descent (SGD) instead of Adam leads to better generalisation \citep{sgd_v_adam}. Furthermore, sharpness-aware minimisation (SAM) \citep{sam} has been proposed to avoid sharp local minima, improving generalisation. The first variation of \citep{robust} uses SAM and SGD instead of Adam. The learning rate for SGD is set to 0.1, with a momentum of 0.9. Adaptive SAM is used with $\rho=2$.
    
    The keypoint layout, i.e. the number and distribution of the keypoints in the field template, affects performance \citep{robust}. Another variation, therefore, uses the grass band keypoints \citep{grass_band}. To investigate the effect of uniform versus non-uniform keypoint layouts, where the total number of keypoints remains constant, yet another variation considers an increased number of uniform keypoints, such that the total matches that of the grass band keypoints (i.e. 147 keypoints with a uniform spatial distribution).
    
    \subsection{Introduction to evaluation metrics}
    \label{sec:metrics}
    Following \citep{robust, kpsfr, real_time_pose}, the mean and median of various evaluation metrics are reported in section \ref{sec:results_discuss}. These metrics are briefly explained, followed by more detailed explanations in the following subsections.

    \subsubsection{Homography evaluation metrics}
        Two types of IoU metrics evaluate the homographic projections of the video frame and the field template, respectively. Additionally, the projection error of randomly sampled points in the video frame projected onto the field template and the re-projection error of the field template keypoints into the video frame are reported. Re-projection refers to transforming a coordinate in the field template to a point in the video frame using the ground truth or estimated (predicted) homography. Projection refers to the inverse of this transformation. 

    \subsubsection{Keypoint measurement metrics}
        The normalised root-mean-square errors (NRMSE) for keypoint coordinates in the x and y image dimensions are reported to evaluate image keypoint position estimates. Furthermore, precision and recall are reported for a given distance threshold, while the mean average precision (mAP) is used to evaluate keypoint detections over a range of distance thresholds.

    \subsection{Evaluation metrics}
        \subsubsection{Intersection over union}
        \label{sec:IoU}
            The first type of IoU considered, $\text{IoU}_{\text{entire}}$, is obtained by re-projecting the field template mask -- i.e. the rectangle that represents the soccer field -- using the ground truth homography. This re-projection is then projected using the predicted homography. The $\text{IoU}_{\text{entire}}$ is equal to the area of intersection of this projected polygon and the field template polygon, divided by the area of their union. I.e. if the estimated homography performs the same mapping as that of the ground truth over the entire field, $\text{IoU}_{\text{entire}}=1$. This definition of $\text{IoU}_{\text{entire}}$ is consistent with \citep{robust, real_time_pose}. However, the $\text{IoU}_{\text{entire}}$ is calculated incorrectly in \citep{kpsfr}. Instead of using the field template mask, \citep{kpsfr} projects the binary mask representing the image area onto the field template using the ground truth homography. The resulting polygon is then re-projected to image coordinates using the predicted homography. The $\text{IoU}_{\text{entire}}$ is obtained as the area of intersection of the re-projected polygon and the binary mask representing the image area, divided by the area of their union\footnote{\url{https://github.com/ericsujw/KpSFR/blob/main/metrics.py}.}.

            The second type of IoU, $\text{IoU}_{\text{part}}$, considers only the part of the field which is visible in the video frame. The ground truth homography projects the binary mask representing the image area. The predicted homography also performs this projection. Each of these projections results in a polygon in the field template. The $\text{IoU}_{\text{part}}$ is equal to the area of intersection of these polygons divided by their area of union.

            A point of concern in \citep{real_time_pose} is that $\text{IoU}_{\text{part}}$ does not take the mapping accuracy outside of the visible field into account. Nevertheless, it was pointed out in \citep{robust} that the ground truth homography is determined only from the visible part of the field. Therefore, $\text{IoU}_{\text{entire}}$ is not necessarily reliable since the ground truth mapping is not guaranteed to be accurate outside of the visible field. These concerns are valid, and annotations are never perfect. However, with the re-annotation of the WC14 and TSWC datasets, resulting in the CARWC dataset, it is believed that both of these metrics are more reliable.

        \subsubsection{Projection error}
        \label{sec:proj}
            The projection error is calculated similarly to \citep{kpsfr, robust}. First, 2500 image points are randomly sampled from a uniform distribution defined over the part of the image where the soccer field is visible. This image portion is determined by the re-projection of the field template mask using the ground truth homography. The projection error in meters is then calculated as the average pair-wise distance between the projections of these points using the predicted homography and their projections using the ground truth homography. However, whereas \citep{robust, kpsfr} assumed field dimensions of $100 \times 60$ m, dimensions of $105 \times 68$ m are used instead (unless otherwise specified) -- which more accurately reflect FIFA regulations \citep{fifa}.

        \subsubsection{Re-projection error}
        \label{sec:reproj}
            Re-projection error is the average pair-wise distance between the field template keypoints re-projected in the image using the ground truth homography and those re-projected using the predicted homography, normalised by the image height \citep{kpsfr, robust}.

        \subsubsection{Normalised root-mean-square error}
        \label{sec:rmse}
            To investigate the effect of keypoint filtering on keypoint position estimates, the NRMSE is used:
                \begin{equation*}
                    \operatorname{NRMSE}=\frac{1}{Z\sqrt{L}}\sqrt{\sum\limits_{l=1}^L\left(x^{I,j}_l-\hat{x}^{I,j}_l\right)^2},
                \end{equation*}
            where $L$ is the total number of measured keypoints which correspond to the ground truth, $x^{I,j}_l$ and $\hat{x}^{I,j}_l$ are the corresponding ground truth and estimated keypoint positions, respectively, in either the x- or y-dimension. Finally, $Z$ is the image width or height corresponding to the computation of the NRMSE in the x- or y-dimension.

        \subsubsection{Precision, recall and mean-average precision}
        \label{sec:prmap}
            Precision is the ratio of true positive detections to the number of predicted detections, while recall is the ratio of true positive detections to the number of ground truth detections. Following \citep{kpsfr}, a keypoint is considered a true positive if it is within a distance of 5 pixels to the ground truth position in the predicted image space ($320\times 180$), which is equivalent to 20 pixels in the actual image space ($1280\times 720$) or $\sim2.78\%$ of the image height. Additionally, the average precision (AP) is evaluated at 5, 10, 15 and 20-pixel thresholds (in the actual image space):
                \begin{equation*}
                    \operatorname{AP}=\sum\limits_n\left(R_n-R_{n-1}\right)P_n,
                \end{equation*} 
            where $R_n$ and $P_n$ are the recall and precision at the $n^{th}$ threshold. The mAP is then obtained as the mean AP.

\section{Results and discussion}
\label{sec:results_discuss}
    The following evaluation metrics are obtained by performing inference on the CARWC test set unless specified otherwise. 

    \subsection{Baseline results}
    \label{sec:baseline_results}
        The results of the baseline methods are presented in Table \ref{tab:baselines_homo} and Table \ref{tab:baselines_kps}, where they are marked with a cross in the BHITK column. Table \ref{tab:baselines_homo} shows the homography evaluation metrics, and Table \ref{tab:baselines_kps} shows the keypoint detection and measurement metrics.
    
        The online refinement algorithm using two loss functions proposed by \citep{robust} did not improve the results. This may be because the self-verification step, which must fail for the online refinement to occur, is consistently successful. In other words, according to the criteria of the online refinement algorithm, the estimated homography is sufficient for most of the CARWC test set. This is consistent with the results presented in \citep{robust}, where this refinement algorithm had an insignificant impact on the WC14 dataset.
        
        The use of SAM and SGD improves every metric when compared to the network trained with Adam (except $\text{IoU}_{\text{entire}}$), thus confirming their positive effect on the generalisation of keypoint detection. With SAM and SGD, recall increases from 90.59\% to 95.10\%, and mAP increases from 61.70\% to 68.58\%. The NRMSE, projection error and re-projection error are also lowered significantly. Despite the improvements to the keypoint detection metrics and $\text{IoU}_{\text{part}}$, the performance of $\text{IoU}_{\text{entire}}$ is degraded compared to the network trained with Adam. This is because RANSAC \citep{ransac} only considers keypoints visible in the current frame, with no mechanism to maintain consistency in the homography between subsequent frames. Thus, it is possible for the out-of-frame mapping to be inconsistent while the within-frame mapping improves. Another explanation could be that the detection model prefers specific keypoints over others. This preference could be due to similarities between the preferred and training data keypoints, while the undetected or less preferred keypoints may be dissimilar. Thus, it is possible that keypoints which may have been essential to obtain an accurate $\text{IoU}_{\text{entire}}$ are missed. However, this explanation is less likely since the keypoint detection metrics (specifically recall) are relatively high.
        
        Adding more uniform keypoints to the field template slightly improves the $\text{IoU}_{\text{part}}$, projection and re-projection metrics but degrades the keypoint detection metrics and $\text{IoU}_{\text{entire}}$. The increased dimensionality of the output may explain the degradation in detection metrics. Nevertheless, this is not sufficient to negate the positive effect of having an increased number of detected keypoints on $\text{IoU}_{\text{part}}$.
        
        The use of grass band keypoints results in the highest precision, lowest re-projection error, and highest mean $\text{IoU}_{\text{part}}$ obtained amongst all the baseline methods. However, it may be concluded from the more considerable difference between the mean and median $\text{IoU}_{\text{entire}}$ (3.61\%) that there are more outlier homography estimates when the entire field is taken into consideration. \mbox{Although} detections are more precise, possibly due to keypoints being placed in more semantically meaningful locations, recall and mAP are lower than for uniform keypoints.
        
        Interestingly, KpSFR \citep{kpsfr} obtains the worst projection error and fairs quite poorly when considering the re-projection and keypoint detection metrics. Nevertheless, it achieves the highest $\text{IoU}_{\text{entire}}$ metrics by a comfortable margin. It slightly improves upon the best median $\text{IoU}_{\text{part}}$ achieved by the variations of \citep{robust}.

    \subsection{Results with the proposed method}
    \label{sec:proposed_results}
        The results of augmenting the baseline methods with the proposed approach are in Table \ref{tab:baselines_homo} and Table \ref{tab:baselines_kps}, marked with a checkmark in the BHITK column. For each distinct variation of \citep{robust}, all evaluation metrics improve using BHITK. The $\text{IoU}_{\text{entire}}$ benefits significantly from homography filtering: the large difference between the mean and median of $\text{IoU}_{\text{entire}}$ for the variation which uses grass band keypoints has been reduced from 3.61\% to 1.99\% (i.e. there are fewer outlier homography estimates compared to the baseline). Furthermore, the mean $\text{IoU}_{\text{entire}}$ increased by 7.04\% for this variation. Relative to the unaugmented mean $\text{IoU}_{\text{entire}}$, this is an improvement of 8.32\%. Significant improvements of the $\text{IoU}_{\text{entire}}$ metrics are also seen for the other baseline methods. While $\text{IoU}_{\text{part}}$ also improves with BHITK, these improvements are less striking since the $\text{IoU}_{\text{part}}$ obtained without BHITK is already relatively high. The best of the unaugmented mean and median projection errors were reduced by 7 cm and 6 cm, respectively, with BHITK. Similarly, the best re-projection errors were improved by 0.15\% and 0.14\% (when compared to the best of these metrics obtained with BHITK over all the experiments). These improvements may seem marginal, but the percentage improvement relative to the results obtained with the unaugmented methods is significant, as shown in Table \ref{tab:results_summary}.

        With BHITK, the best mAP increased from 68.58\% to 69.97\%, the best recall from 95.10\% to 95.36\% and the best precision from 96.44\% to 96.95\%. There are also minor improvements to the best of the NRMSE metrics. This shows that the keypoint positions are effectively refined, contributing to the homography filter's effectiveness. However, the low mAP metrics relative to the generally high precision and recall suggest that keypoint identification still suffers at lower thresholds.

        Augmenting \citep{robust}, with no variations, with BHITK improves nearly all homography evaluation metrics over those achieved by KpSFR \citep{kpsfr}, except for the median re-projection error where the difference is only 0.02\% (a relative degradation of 2.82\%). This is despite having a much lower mAP, recall and slightly higher NRMSE in the x dimension. The improvement is especially significant considering the increased parameter count (73 million) and inference time (1.5 frames per second) of KpSFR compared to \citep{robust} (42 million and 50 frames per second, respectively). Thus, BHITK enables a less sophisticated and less computationally expensive method to outperform the state-of-the-art in most homography evaluation metrics. Furthermore, these improvements are obtained without necessarily altering the existing method. Finally, BHITK may enable performance that is not attainable by considering such alterations, as is shown by the fact that each distinct variation improved significantly using BHITK. \begin{table*}[hbt!]
        \caption{A comparison of the homography evaluation performance of the baseline detection methods with and without BHITK on the time-series part of the CARWC dataset. Models augmented with BHITK are marked with a checkmark in the BHITK column.}
        \centering
            \newcommand{\methodwidth}{4cm}
            \begin{tabular}{>{\rowmac}c >{\rowmac}c >{\rowmac}c>{\rowmac}c >{\rowmac}c>{\rowmac}c >{\rowmac}c>{\rowmac}c >{\rowmac}c>{\rowmac}c<{\clearrow}}
                \toprule \multirow{2}{\methodwidth}{Method} & \multirow{2}{*}{BHITK} & \multicolumn{2}{c}{$\text{IoU}_{\text {entire }}(\%) \uparrow$} & \multicolumn{2}{c}{$\text{IoU}_{\text{part}}(\%) \uparrow$} & \multicolumn{2}{c}{ Proj. (meter) $\downarrow$} & \multicolumn{2}{c}{ Re-Proj. (\%)$\downarrow$}\\
                
                \cmidrule{3-10} & & mean & median & mean & median & mean & median & mean & median\\ 
                
                \specialrule{1.25pt}{\toppad}{\botpad} \multirow{1}{\methodwidth}{\cite{robust} with online refinement} & \text{\sffamily X} & 86.76 & 89.67 & 98.18 & 98.43 & 0.38 & 0.35 & 0.89 & 0.78\\
    
                \specialrule{1.25pt}{\toppad}{\botpad} \multirow{2}{\methodwidth}{\cite{robust}} & \text{\sffamily X} & 86.79 & 89.67 & 98.19 & 98.43 & 0.37 & 0.35 & 0.88 & 0.78\\

                \cmidrule{2-10} & \checkmark & \setrow{\bfseries} 90.48 & 92.14 & 98.63 & 98.88 & 0.34 & 0.33 & 0.77 & 0.73\\
                
                \specialrule{1.25pt}{\toppad}{\botpad} \multirow{2}{\methodwidth}{\cite{robust} + SAM + SGD} & \text{\sffamily X} & 85.77 & 87.93 & 98.37 & 98.64 & 0.32 & 0.28 & 0.77 & 0.71\\

                \cmidrule{2-10} & \checkmark & \setrow{\bfseries} 92.29 & 93.68 & 98.87 & 99.00 & 0.25 & 0.22 & 0.59 & 0.57\\
                
                \specialrule{1.25pt}{\toppad}{\botpad} \multirow{2}{\methodwidth}{\cite{robust} + SAM + SGD + more KPs} & \text{\sffamily X} & 85.08 & 87.31 & 98.48 & 98.67 & 0.30 & 0.28 & 0.76 & 0.68\\

                \cmidrule{2-10} & \checkmark & \setrow{\bfseries} 91.33 & 92.99 & 98.89 & 99.06 & 0.26 & 0.23 & 0.61 & 0.52\\
                
                \specialrule{1.25pt}{\toppad}{\botpad} \multirow{2}{\methodwidth}{\cite{robust} + SAM + SGD + grass band KPs} & \text{\sffamily X} & 84.65 & 88.26 & 98.52 & 98.66 & 0.30 & 0.28 & 0.70 & 0.66\\

                \cmidrule{2-10} & \checkmark & \setrow{\bfseries} 91.69 & 93.68 & 98.94 & 99.08 & 0.23 & 0.23 & 0.55 & 0.53\\
                
                \specialrule{1.25pt}{\toppad}{\botpad} \multirow{1}{\methodwidth}{KpSFR \citep{kpsfr}} & \text{\sffamily X} & 89.52 & 91.40 & 98.36 & 98.68 & 0.42 & 0.37 & 0.83 & 0.71\\
                
                \bottomrule
            \end{tabular}
        \label{tab:baselines_homo}
        \end{table*} \begin{table*}[hbt!]
        \caption{A comparison of the keypoint detection and measurement performance of the baseline detection methods with and without BHITK on the time-series part of the CARWC dataset. Models augmented with BHITK are marked with a checkmark in the BHITK column.}
        \centering
            \newcommand{\methodwidth}{5cm}
            \begin{tabular}{>{\rowmac}c >{\rowmac}c >{\rowmac}c>{\rowmac}c >{\rowmac}c>{\rowmac}c>{\rowmac}c<{\clearrow}}
                \toprule \multirow{2}{*}{Method} & \multirow{2}{*}{BHITK} & \multicolumn{2}{c}{NRMSE (\%)$\downarrow$} & \multirow{2}{0.6cm}{P(\%)$\uparrow$} & \multirow{2}{0.6cm}{R(\&)$\uparrow$} & \multirow{2}{1cm}{mAP(\%)$\uparrow$}\\
                
                 \cmidrule{3-4}& & y & x & & & \\ 
                
                \specialrule{1.25pt}{\toppad}{\botpad} \multirow{1}{*}{\cite{robust} with online refinement} & \text{\sffamily X} & 0.65 & 0.71 & 94.98 & 90.59 & 61.70\\
    
                \specialrule{1.25pt}{\toppad}{\botpad} \multirow{2}{*}{\cite{robust}} & \text{\sffamily X} & 0.65 & 0.71 & 94.98 & 90.59 & 61.70\\

                 & \checkmark & \setrow{\bfseries} 0.63 & 0.68 & 95.49 & 91.08 & 62.26\\
                
                \specialrule{1.25pt}{\toppad}{\botpad} \multirow{2}{*}{\cite{robust} + SAM + SGD} & \text{\sffamily X} & 0.53 & 0.57 & 96.14 & 95.10 & 68.58\\

                 & \checkmark & \setrow{\bfseries} 0.50 & 0.55 & 96.42 & 95.36 & 69.97\\
                
                \specialrule{1.25pt}{\toppad}{\botpad} \multirow{2}{*}{\cite{robust} + SAM + SGD + more KPs} & \text{\sffamily X} & 0.59 & 0.58 & 95.95 & 94.96 & 66.77\\

                 & \checkmark & \setrow{\bfseries} 0.55 & 0.56 & 96.23 & 95.24 & 67.87\\
                
                \specialrule{1.25pt}{\toppad}{\botpad} \multirow{2}{*}{\cite{robust} + SAM + SGD + grass band KPs} & \text{\sffamily X} & 0.63 & 0.58 & 96.44 & 94.86 & 65.39\\

                 & \checkmark & \setrow{\bfseries} 0.58 & 0.55 & 96.95 & 95.36 & 66.37\\
                
                \specialrule{1.25pt}{\toppad}{\botpad} \multirow{1}{*}{KpSFR \citep{kpsfr}} & \text{\sffamily X} & 0.67 & 0.66 & 95.13 & 93.28 & 66.58\\
                
                \bottomrule
            \end{tabular}
        \label{tab:baselines_kps}
        \end{table*} \begin{table*}[hbt!]
        \caption{The percentage improvement of the best performance metrics obtained with BHITK, relative to those obtained without BHITK.}
        \centering
        \setlength{\tabcolsep}{5pt}
            \begin{tabular}{>{\rowmac}c>{\rowmac}c >{\rowmac}c>{\rowmac}c >{\rowmac}c>{\rowmac}c >{\rowmac}c>{\rowmac}c >{\rowmac}c>{\rowmac}c >{\rowmac}c>{\rowmac}c>{\rowmac}c<{\clearrow}}
                \toprule \multicolumn{2}{c}{$\text{IoU}_{\text {entire }}$} & \multicolumn{2}{c}{$\text{IoU}_{\text{part}}$} & \multicolumn{2}{c}{ Proj. (meter)} & \multicolumn{2}{c}{ Re-Proj.} & \multicolumn{2}{c}{NRMSE} & \multirow{2}{*}{P} & \multirow{2}{*}{R} & \multirow{2}{*}{mAP}\\
                
                \cmidrule{1-10} mean & median & mean & median & mean & median & mean & median & y & x & & & \\ 

                \midrule \setrow{\bfseries} 3.09\% & 2.49\% & 0.43\% & 0.41\% & 23.33\% & 21.43\% & 21.43\% & 21.21\% & 5.66\% & 3.51\% & 0.53\% & 0.27\% & 2.03\%\\

                \bottomrule
            \end{tabular}
        \label{tab:results_summary}
        \end{table*}

    \subsection{Results on TSWC}
    \label{sec:tswc_results}
        The proposed method is applied to a variation of \citep{robust} without dense feature regression, as implemented in \citep{kpsfr}, to compare results on the TSWC dataset. Table \ref{tab:ekf_tswc} shows that the improvement afforded by the proposed method is consistent across the different datasets. The simplified \citep{robust} network with BHITK outperforms or achieves similar homography metrics to more computationally expensive methods, such as KpSFR \citep{kpsfr} and the method of \citep{synthetic}. These improvements are achieved despite a much lower recall than that of KpSFR -- similar to the improvement of \citep{robust} over KpSFR despite lower recall in \ref{sec:proposed_results}. Most keypoint detection metrics also do not show a significant improvement. Thus, it may be concluded that the homography filter plays the most significant role in the proposed method. \begin{table*}[hbt!]
        \caption{A comparison of BHITK with past methods on the TSWC dataset.}
        \centering
        \setlength{\tabcolsep}{5pt}
            \begin{tabular}{m{40pt} >{\rowmac}c>{\rowmac}c >{\rowmac}c>{\rowmac}c >{\rowmac}c>{\rowmac}c >{\rowmac}c>{\rowmac}c >{\rowmac}c>{\rowmac}c >{\rowmac}c >{\rowmac}c >{\rowmac}c<{\clearrow}}
                \toprule \multirow{2}{*}{ Method } & \multicolumn{2}{c}{$\text{IoU}_{\text {entire }}(\%) \uparrow$} & \multicolumn{2}{c}{$\text{IoU}_{\text{part}}(\%) \uparrow$} & \multicolumn{2}{c}{ Proj. (meter) $\downarrow$} & \multicolumn{2}{c}{ Re-Proj. (\%)$\downarrow$} & \multicolumn{2}{m{1.1cm}}{NRMSE (\%) $\downarrow$} & \multirow{2}{0.6cm}{P(\%)$\uparrow$} & \multirow{2}{0.6cm}{R(\%)$\uparrow$} & \multirow{2}{1cm}{mAP(\%)$\uparrow$}\\
                \cmidrule{2-11} & mean & median & mean & median & mean & median & mean & median & y & x & & & \\
    
                \midrule \cite{synthetic} as reported in \citep{kpsfr} & 90.7$^*$ & 94.1$^*$ & 96.8 & 97.4 & 0.54$^\dagger$ & 0.38$^\dagger$ & 1.6 & 1.3 & -- & -- & -- & -- & --\\
                
                \midrule \cite{robust} as in \citep{kpsfr} & 92.5$^*$ & 94.2$^*$ & 97.4 & 97.9 & 0.43$^\dagger$ & 0.37$^\dagger$ & 1.1 & 1.0 & 0.66 & 0.78 & 94.96 & 83.23 & 56.87\\
                
                \midrule KpSFR \citep{kpsfr} & 94.8$^*$ & 95.4$^*$ & 98.1 & 98.2 & 0.36$^\dagger$ & 0.33$^\dagger$ & 0.9 & 0.8 & -- & -- & -- & 87 & --\\
    
                \midrule \cite{robust} as in \cite{kpsfr} + \textbf{BHITK} & \setrow{\bfseries} 94.8$^*$ & 95.8$^*$ & 97.9 & 98.3 & 0.36$^\dagger$ & 0.32$^\dagger$ & 0.9 & 0.8 & 0.62 & 0.76 & 95.06 & 83.32 & 56.92\\
                \bottomrule
                \multicolumn{14}{l}{\footnotesize $^*$The $\text{IoU}_{\text {entire }}$ metric is calculated as in \citep{kpsfr}.}
                \\
                \multicolumn{14}{l}{\footnotesize $^\dagger$The projection error is calculated using field template dimensions of 100 $\times$ 60 meters, as in \citep{kpsfr}.}
            \end{tabular}
        \label{tab:ekf_tswc}
        \end{table*}

    \subsection{Qualitative evaluations}
    \label{sec:qualitative_eval}
\begin{figure*}[hbt!]
                \centering
                \caption{This figure depicts both the re-projected keypoints within the image and the image projected onto the field template. The projection and re-projection in each sub-figure utilise distinct homography estimates: one derived from \citep{robust}'s method and the other obtained by augmenting this same network with the proposed method (BHITK). The sub-figures represent the same frame from the same test video. The red circles represent the keypoints re-projected using the predicted homography, and the green circles represent the keypoints re-projected using the ground truth homography.}
                    \begin{subfigure}[]{\textwidth}
                        \centering
                        \includegraphics[width=\textwidth]{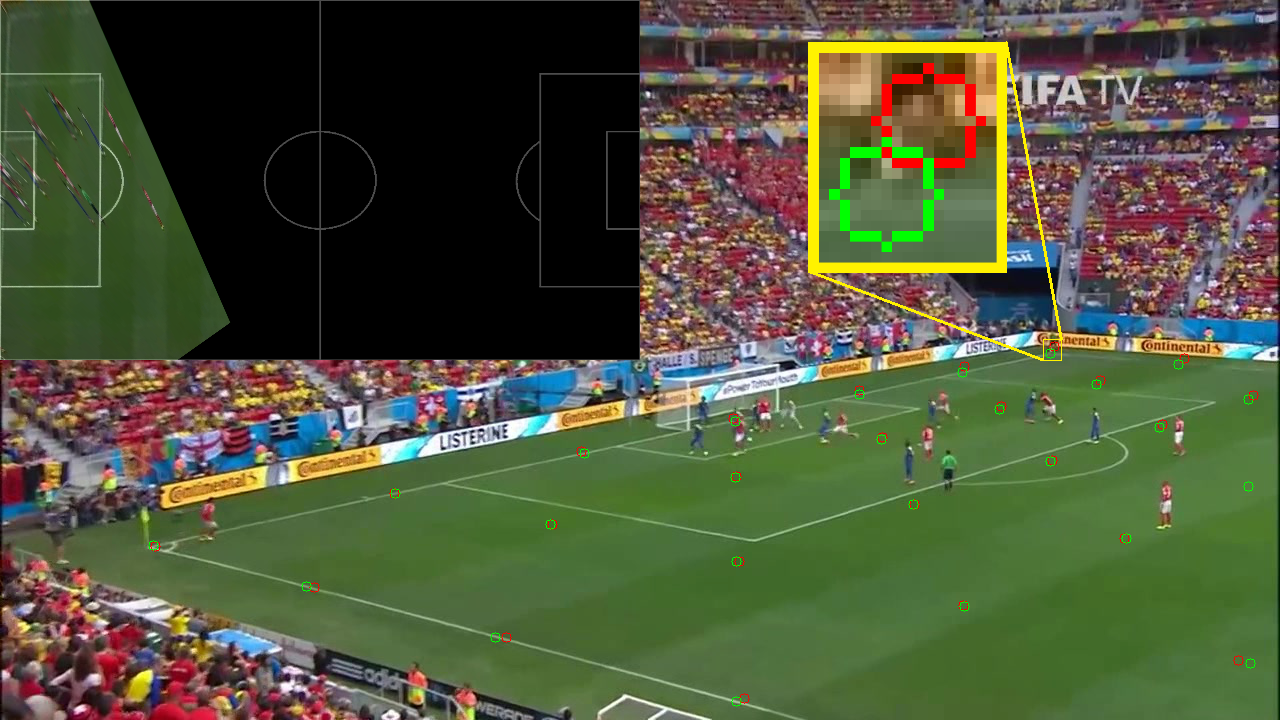}
                        \caption{Projection and re-projection using the homography estimate obtained with \citep{robust}.}
                        \label{fig:unfilt_adam}
                    \end{subfigure}
                    \hfill
                    \begin{subfigure}[]{\textwidth}
                        \centering
                        \includegraphics[width=\textwidth]{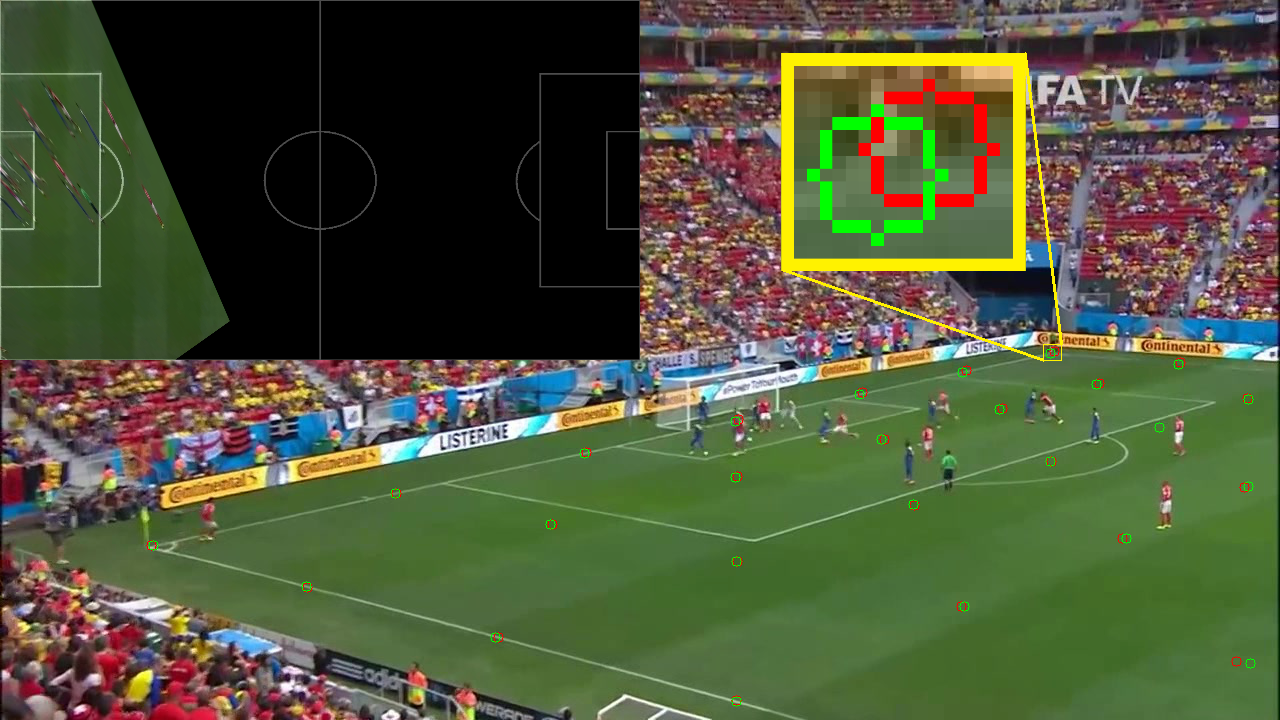}
                        \caption{Projection and re-projection using the homography estimate obtained with \citep{robust} + BHITK.}
                        \label{fig:filt_adam}
                    \end{subfigure}
                \label{fig:unfilt_filt_comp_adam}
            \end{figure*}

\begin{figure*}[hbt!]
                \centering
                \caption{This figure depicts both the re-projected keypoints within the image and the image projected onto the field template. The projection and re-projection in each sub-figure utilise distinct homography estimates: one derived from a variant of \citep{robust}'s method, which is trained with SAM, SGD and 147 keypoints, and the other obtained by augmenting this same network with the proposed method (BHITK). The sub-figures represent the same frame from the same test video. The red circles represent the keypoints re-projected using the predicted homography, and the green circles represent the keypoints re-projected using the ground truth homography.}
                    \begin{subfigure}[]{\textwidth}
                        \centering
                        \includegraphics[width=\textwidth]{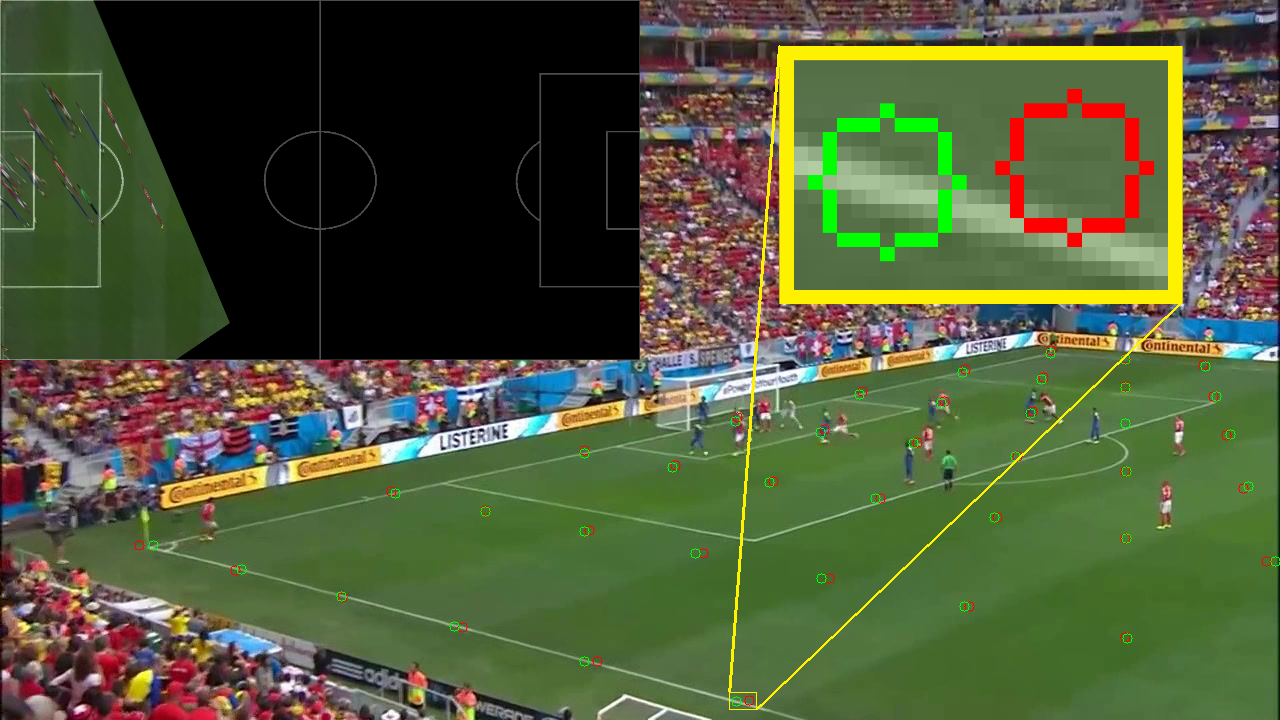}
                        \caption{Projection and re-projection using the homography estimate obtained with \citep{robust} + SAM + SGD + more KPs.}
                        \label{fig:unfilt}
                    \end{subfigure}
                    \hfill
                    \begin{subfigure}[]{\textwidth}
                        \centering
                        \includegraphics[width=\textwidth]{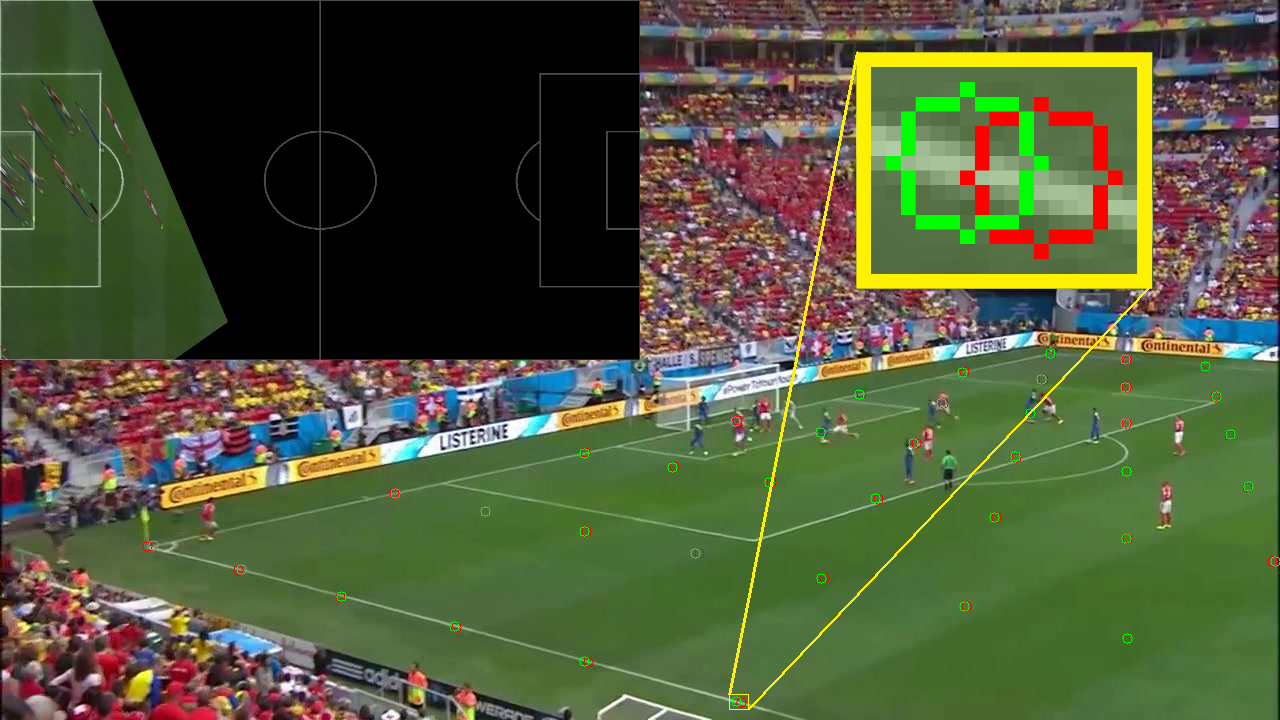}
                        \caption{Projection and re-projection using the homography estimate obtained with \citep{robust} + SAM + SGD + more KPs + BHITK.}
                        \label{fig:filt}
                    \end{subfigure}
                \label{fig:unfilt_filt_comp}
            \end{figure*}

        Qualitative comparisons are shown in Fig. \ref{fig:unfilt_filt_comp_adam} and Fig. \ref{fig:unfilt_filt_comp}. In each figure, the homography estimate predicted for the same image is used to re-project the field template keypoints into the image and project the image onto the field template. In each case, the first sub-figure represents the results obtained without using BHITK, while the second shows the results using BHITK. Specifically, Fig. \ref{fig:unfilt_adam} shows the results obtained with the homography estimate obtained from the method of \citep{robust}, while Fig. \ref{fig:filt_adam} shows the results obtained when the same method is augmented with BHITK. Similarly, Fig. \ref{fig:unfilt} shows the results obtained with the variation of \citep{robust} trained with SAM, SGD and 147 keypoints, while Fig. \ref{fig:filt} shows the results when this variation is augmented with BHITK.
        
        In Fig. \ref{fig:unfilt_filt_comp_adam} and Fig. \ref{fig:unfilt_filt_comp}, the re-projected keypoints are closer to their ground truth positions when BHITK is employed. Furthermore, the projected field lines align more closely with those of the field template when using BHITK.
             

\section{Conclusion}
\label{sec:Conclusion}
    Exploiting the temporally consistent nature of homographic projections in a Bayesian framework is shown to be beneficial. The proposed approach effectively enforces temporal consistency between subsequent homography estimates through an affine transformation. As long as the underlying keypoint detection method satisfies the standard Kalman filter assumptions (i.e. approximately zero-mean Gaussian-distributed measurement noise), homography filtering from tracked keypoints is shown to be effective. When augmented with the proposed method, the overall weakest-performing baseline method outperforms the state-of-the-art, which is much more computationally expensive, in all but one of the homography evaluation metrics (median re-projection error, where the difference is only 0.02\%). Furthermore, all baseline evaluation metrics improve when the baseline methods are augmented with BHITK. Thus, the method will likely improve the performance of several existing keypoint detection methods. Finally, the annotations of the WorldCup and TS-WorldCup datasets are refined and released along with a custom homography annotation tool as the CARWC dataset.

\section*{CRediT authorship contribution statement}
    \textbf{Paul Claasen}: Conceptualization, Methodology, Software, Validation, Formal analysis, Investigation, Resources, Data Curation, Writing - Original Draft, Visualization. \textbf{Pieter de Villiers}: Conceptualization, Writing - Review \& Editing, Supervision, Project administration, Funding acquisition.


\section*{Data availability}
    The used datasets are already publicly available.

\section*{Acknowledgements}
    This work was supported by the MultiChoice Chair in Machine Learning and the MultiChoice Group.
    



 \bibliographystyle{elsarticle-num-names} 
 \bibliography{references.bib}





\end{document}